\documentclass[11pt]{article}

\usepackage[final]{acl}

\usepackage{times}
\usepackage{latexsym}

\usepackage[T1]{fontenc}

\usepackage[utf8]{inputenc}

\usepackage{microtype}

\usepackage{inconsolata}
\usepackage{graphicx}
\usepackage{amsmath,hyperref}
\usepackage{amsfonts}
\usepackage{enumitem}
\usepackage[table]{xcolor}
\usepackage{booktabs}
\usepackage{multirow}
\usepackage{stfloats}
\usepackage{makecell}
\usepackage{tabularx,booktabs}
\usepackage{subcaption}
\usepackage[most]{tcolorbox}
\usepackage{xcolor}
\usepackage{listings}
\usepackage{placeins}
\newcolumntype{Y}{>{\ttfamily\footnotesize\raggedright\arraybackslash}X} 
\newcolumntype{R}{>{\footnotesize\raggedright\arraybackslash}p{2.2cm}} 
\usepackage{xcolor}

\newtcolorbox{promptbox}{
  colback=gray!10,    
  colframe=gray!40,   
  boxrule=0.3pt,      
  arc=2pt,            
  left=4pt, right=4pt, top=4pt, bottom=4pt
}

\title{TLRD: Teaching LLMs to Reason over Tabular Data with Tri-Level Rationale Distillation}

\author{
 \textbf{Tianyuan Liang\textsuperscript{1 *}},
 \textbf{Xuwei Tan\textsuperscript{1 * }\thanks{Project lead.}},
 \textbf{Lei Shi\textsuperscript{1}},
 \textbf{Junsheng Zhong\textsuperscript{1}},
 \textbf{Ziyu Hu\textsuperscript{2}},
 \\
 \textbf{Tian Xie\textsuperscript{1}},
 \textbf{Zhiqun Zuo\textsuperscript{1}},
 \textbf{Xiaodong Yu\textsuperscript{2}},
 \textbf{Xueru Zhang\textsuperscript{1}}
\\
 \textsuperscript{1}The Ohio State University \quad
 \textsuperscript{2}Stevens Institute of Technology
\\
 \small
 \textsuperscript{*}Equal contribution.
}

\begin{document}
\maketitle
\begin{abstract}

Tabular data is a primary medium for storing real-world information, driving many industrial applications of machine learning. Traditional predictors achieve strong predictive performance but do not provide readable, case-specific explanations essential for decision-making. Large Language Models (LLMs) can naturally bridge this gap by generating predictions alongside explanations. However, dataset-specific patterns, such as feature distributions and interactions, make tabular data difficult for LLMs to understand and reason over, while label-only fine-tuning improves performance at the cost of catastrophic forgetting. To address this problem, we propose Tri-Level Rationale Distillation (TLRD), a framework that converts label-only tabular datasets into structured rationale supervision for LLMs. TLRD uses a high-capacity teacher to synthesize a rationale corpus grounded in three complementary levels of evidence: instance-level feature, dataset-level distributional context, and comparison-level retrieved neighbors, then distills the rationale into student LLMs, enabling zero-overhead prediction and grounded explanation from raw features only. Experiments on multiple domain datasets show that TLRD significantly closes the performance gap between LLMs and state-of-the-art tree ensembles while producing grounded and readable explanations, offering a valuable reference for high-stakes decision-making.

\end{abstract}

\section{Introduction} 
\begin{figure}[t]
    \centering
    \includegraphics[width=\columnwidth]{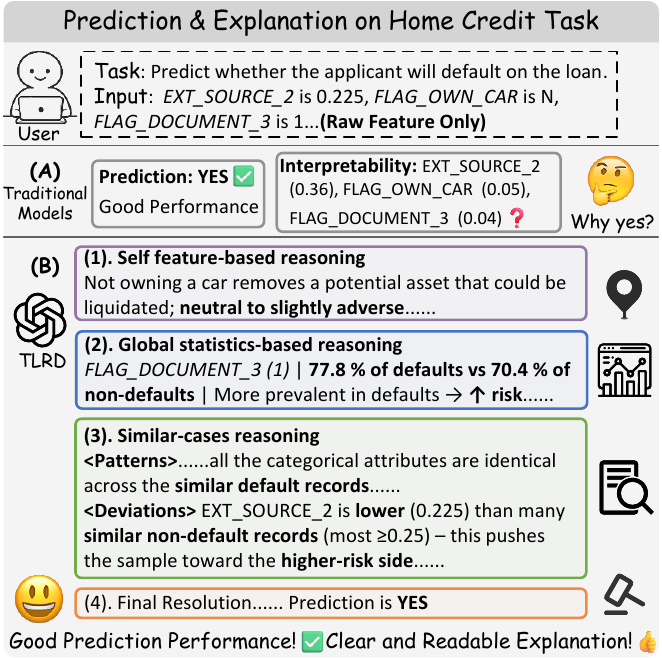}
    \caption{Illustration of a TLRD deployment. Given raw features, the student model generates tri-level rationales.}
    \vspace{-0.1cm}
    \label{fig:example}
\end{figure}

Many industrial decision-making tasks rely on tabular data, including credit scoring, healthcare, supply-chain optimization, and user profiling. In these decision-critical settings, tree-based ensembles such as XGBoost \citep{chen2016xgboost} and CatBoost \citep{prokhorenkova2018catboost} remain the dominant methods, often outperforming deep learning methods under realistic tuning and data-size regimes \cite{shwartz2022tabular,grinsztajn2022tree}. Even as foundation models emerge \citep{hollmann2023tabpfn}, practitioners still favor tree-based ensembles for their efficiency and strong performance. However, model predictions cannot directly translate into practical decisions, since real-world decisions are made by stakeholders under task-specific scenarios, with models serving as assistive tools. In practice, stakeholders need concise and readable rationales (\emph{Figure~\ref{fig:example} (B)}) that explain (i) which specific feature values in this case are concerning, and (ii) how unusual they are relative to the broader population. However, tree-based ensembles provide only coarse interpretability signals (e.g., \emph{feature attribution scores in Figure~\ref{fig:example} (A)}) that do not align with operational decision needs. Recent LLM-based methods even claim to outperform strong baselines such as XGBoost \citep{gardner2024large,wang2026boostllmboostinginspiredllmfinetuning}, but they are established under few-shot settings, which are less practical, while optimizing explanation quality remains largely underexplored. Accordingly, our work aims to \textbf{provide complementary decision support for tabular data, offering a practical trade-off between performance and interpretability.}

LLMs naturally bridge the gap between predictive capability and human-readable reasoning. However, in zero-shot or few-shot prediction tasks, LLMs often exhibit weak performance on certain tasks \cite{xie2024finben, pmlr-v206-hegselmann23a,wolff2025well}, especially those requiring precise numerical comparisons or domain-specific feature interactions not encountered during pre-training. A natural response is to fine-tune an LLM directly on labeled tabular examples. However, this introduces a new technical hurdle: the catastrophic forgetting \cite{luo2025empirical} of generative behavior. When supervised solely by categorical labels, the model optimizes strictly for the discriminative mapping, implicitly achieving competitive accuracy while minimizing token length. Consequently, the model becomes biased toward terse label-only outputs and no longer reliably generates free-form explanations. This behavior is also denoted as the explanation collapse \cite{yang2025beyond}. This motivates the central question of our work: \emph{How can we fine-tune a deployable (small) LLM on a label-only tabular dataset to achieve high predictive performance while preserving high-quality, evidence-anchored explanations?}

To address this, we propose \textbf{Tri-Level Rationale Distillation (TLRD)}, a general pipeline that converts label-only tabular datasets into a supervision-rich, structured reasoning corpus. By distilling from a high-capacity teacher model, TLRD synthesizes a structured tri-level rationale report that closely mirrors decision-making workflows:

\begin{itemize}[leftmargin=*, nosep]
  \item \textbf{Instance-level reasoning}: Identifies and isolates the primary feature triggers and local patterns specific to the individual data row, guided by \emph{domain knowledge and commonsense constraints}.
  \item \textbf{Dataset-level reasoning}: Contextualizes specific values against the global feature distribution, anchoring the explanation in \emph{macro-level statistics}.
  \item \textbf{Comparison-level reasoning}: Leverages a few similar cases to distill contrastive insights into \emph{pattern summaries}.

\end{itemize}

During rationale synthesis, TLRD provides the teacher model with a \textbf{Tri-Level Augmented Input} as complementary evidence by extending current row with dataset-level statistics and comparison-level retrieved neighbors, helping the teacher understand feature values not only from the current row, but also through global distributions and local contrastive patterns. Guided by \textbf{Tri-Level Rationale Schema}, the teacher converts this evidence into structured rationales. A compact student model is then fine-tuned on the distilled corpus to predict the label and generate the rationale given only raw features. In this way, dataset context and case-based signals are \emph{indirectly injected} into the student, while the student remains deployable without external statistics or retrieval, avoiding added latency. Our contributions are summarized as follows:

  \begin{itemize}[leftmargin=*,nosep]
    \item 
    \textbf{Tri-Level Evidence to Rationales.}
    We introduce dataset statistics and retrieved neighbors as extra evidence, and verify that they improve base LLMs' prediction. We then formalize a structured schema that guides a teacher to analyze these evidence sources under the ground-truth label and transform them into tri-level rationales.

    \item 
     \textbf{Context-to-Parameter Tabular Rationale Distillation:} 
     We introduce a distillation method that uses dataset statistics and retrieved neighbors only during corpus construction, indirectly injecting dataset-level and case-based knowledge into the student and enabling zero-overhead inference from raw features.

    \item 
    \textbf{Comprehensive Empirical Evaluation:}  

    We evaluate TLRD on multiple decision-critical tabular tasks, showing that TLRD achieves strong predictive performance against strong baselines while producing grounded and readable rationales. We conduct a study of diverse teacher models, student backbones, and self-distillation settings, hoping to bring useful insights.

  \end{itemize}

\section{Related Work}

\subsection{LLMs on Tabular Data}

The adaptation of LLMs for tabular data traditionally relies on serializing structured rows into natural text to bridge the modality gap. Foundational works like TabLLM \citep{pmlr-v206-hegselmann23a} and UniPredict \citep{wang2023unipredict} demonstrated that serialization enables zero-shot and few-shot classification across diverse tabular inputs. However, these methods do not solve LLMs' difficulty in capturing dataset-specific patterns. There are also some methods working on scaling transfer learning on diverse datasets \citep{gardner2024large} and optimizing In-Context Learning (ICL) via advanced prompt engineering \citep{gao2025utilizing}. To address numeric instability and context limits, newer architectures introduce specialized mechanisms, such as TabGemma's \citep{schindler2025tabgemma} numeric canonicalization with retrieval-augmented generation, and highly efficient ICL strategies \citep{wu2025efficient}. Despite these advances, LLMs still struggle to capture dataset-specific patterns, motivating the development of compact, task-specific models that provide grounded explanations.

\begin{figure*}[!t]
    \centering
    \includegraphics[width=0.88\linewidth]{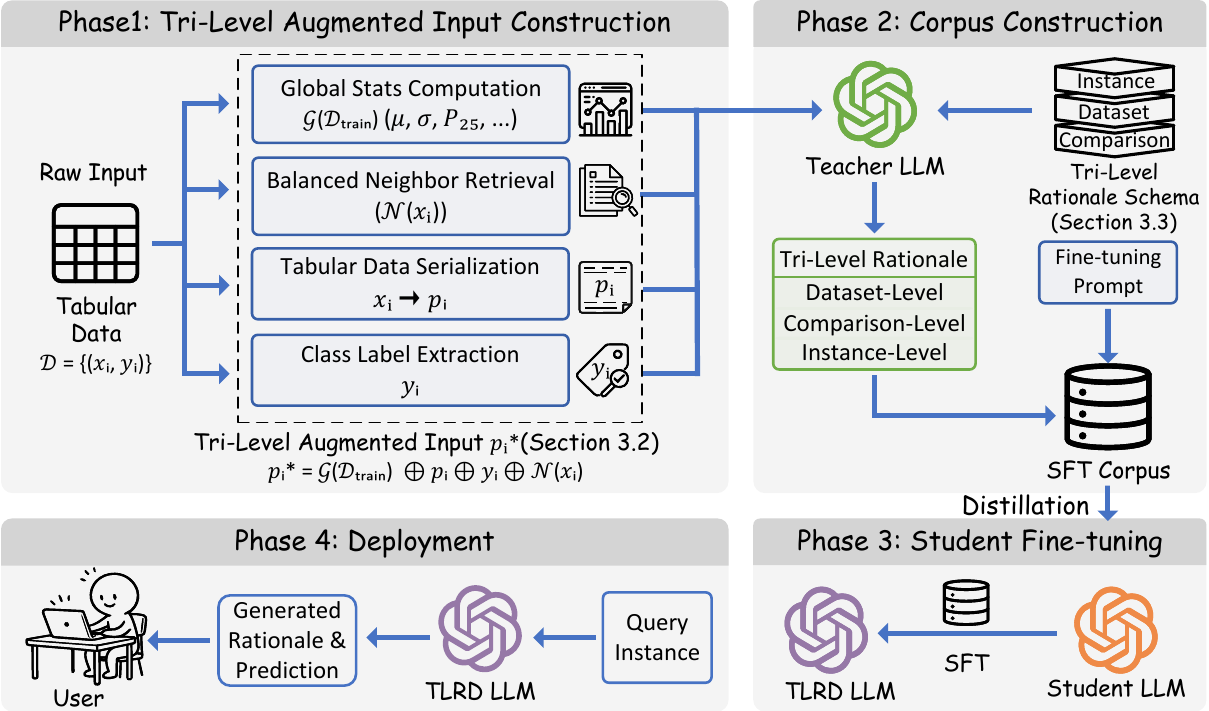}
     \caption{Overview of the TLRD framework. Starting from a label-only tabular dataset, we construct Tri-Level Augmented Inputs for the teacher model. Guided by the Tri-Level Rationale Schema, the teacher generates structured rationales, which are then distilled into a student LLM through supervised fine-tuning. During deployment, the student takes only the raw serialized row as input and outputs both the prediction and the tri-level rationale.}

     \vspace{-0.1cm}
    \label{fig:frame}
\end{figure*}

\subsection{LLM Distillation and Fine-Tuning}

To deploy LLMs efficiently, researchers frequently use knowledge distillation. While traditional distillation transfers only predictive distributions, recent NLP breakthroughs emphasize transferring reasoning abilities from large to small language models and rationale distillation \citep{hsieh2023distilling, magister2023teaching}. However, rationale distillation remains largely underexplored for tabular data, where reasoning often depends on dataset-specific knowledge that cannot be directly acquired from pre-training. Fine-tuning tabular models strictly on labels risks "explanation collapse \citep{yang2025beyond}," where models optimize for accuracy at the expense of generative reasoning faculties. Recent works focus on the performance of few-shot settings \citep{gardner2024large,wang2026boostllmboostinginspiredllmfinetuning}, which are less practical in industrial scenarios where large labeled datasets are usually available. Moreover, their designs do not focus on explanation generation. TLRD bridges these gaps by distilling teacher-generated tri-level rationales based on tri-level evidence and designed schema, enabling the student model to retain grounded, readable reasoning while maintaining strong performance for real-world tabular decision-making settings.

\section{Proposed Method}

\subsection{Problem Formulation}
\label{sec:problem_formulation}

Let $\mathcal{X}$ denote a tabular feature space with both numerical and categorical variables, such that $\mathcal{X}=\mathbb{R}^{d_\text{num}}\times\mathcal{V}^{d_\text{cat}}$, where $d_\text{num}$ and $d_\text{cat}$ are the numbers of numerical and categorical features, respectively, and $\mathcal{V}$ denotes the categorical value space. For a labeled tabular dataset $\mathcal{D}=\{(\mathbf{x}_i,y_i)\}_{i=1}^N$, where $\mathbf{x}_i\in\mathcal{X}$ is the $i$-th instance and $y_i\in\mathcal{Y}$ (e.g., $\mathcal{Y}=\{0,1\}$) is its ground-truth label. Let  $\mathcal{D}_{\text{train}} \subset \mathcal{D}$ denote the training split. Each table row is then converted into a textual prompt $p_i=\mathrm{Serialize}(\mathbf{x}_i)$.

The overall workflow is illustrated in Figure~\ref{fig:frame}. Our goal is to fine-tune an LLM to output both a prediction and a structured, evidence-grounded explanation rationale, i.e., $(y_i, r_i)$ with $r_i \in \mathcal{R}$ following a fixed schema. We seek a mapping that converts label-only dataset into a supervised corpus $\mathcal{C}=\{(p_i,o_i)\mid (\mathbf{x}_i,y_i)\in\mathcal{D}_{\text{train}}\}$, where $o_i=(r_i,y_i)$, by synthesizing schema-compliant rationales from a stronger teacher or via self-distillation.

\subsection{Tri-Level Augmented Input}
\label{sec:tri_level_augmented_input}

To address the limitation that LLMs struggle to understand dataset-specific patterns, we construct a \emph{Tri-Level Augmented Input} that extends the serialized row with global statistics and retrieved neighbors, exposing the teacher to the distributional and contrastive evidence needed for grounded tri-level rationale generation. We first serialize each row $\mathbf{x}\in\mathcal{X}$ in the form of \texttt{<feature> is <value>}, using a fixed order, explicit missing-value placeholders, and append the ground-truth label. We refer to this block as \emph{Current Sample Features} and then augment it with two additional sources of evidence: \emph{Dataset Statistics} and \emph{Retrieved Neighbors}.

\noindent\textbf{Dataset statistics.} To provide global distributional context, we compute and append dataset-level statistics to the teacher prompt. For classification tasks, we compute, for each target class on $\mathcal{D}_{\text{train}}$, numerical-feature statistics including the mean, standard deviation, quartiles (P25, P50, P75), minimum, and maximum, as well as class-conditional category frequencies for categorical features. For regression tasks, we discretize the continuous target into four percentile bins using the 25th, 50th, and 75th percentiles of the target distribution on $\mathcal{D}_{\text{train}}$ as thresholds, and compute the corresponding numerical and categorical feature statistics within each bin. We denote the resulting dataset-level statistical context by $\mathcal{G}(\mathcal{D}_{\text{train}})$.

\noindent\textbf{Retrieved neighbors.}
To provide a contrastive case-based context, we retrieve a label-balanced set of historical instances; the query instance itself is excluded from retrieval. Let $\mathbf{x}_q$ be the query instance, and let $\mathcal{F}_{\text{cat}}$ denote the index set of categorical features. To maintain semantic consistency, we first define a candidate pool $\Omega(\mathbf{x}_q)$ by enforcing categorical feature matches:
\begin{equation*}
\resizebox{0.9\columnwidth}{!}{\ensuremath{
\Omega(\mathbf{x}_q) =
\left\{
\mathbf{x}_i \in \mathcal{D}_{\text{train}}
\mid
\mathbf{x}_i[f] = \mathbf{x}_q[f], \, \forall f \in \mathcal{F}_{\text{cat}}
\right\}.
    }}
\end{equation*}
If this exact-match pool is too small, we progressively relax the categorical matching constraints until a sufficiently large candidate pool is obtained.

Within this structurally aligned pool, we compute a similarity score using standardized numerical features. Let $\mathbf{v}_i$ and $\mathbf{v}_q$ denote the standardized numerical feature vectors of $\mathbf{x}_i$ and $\mathbf{x}_q$, respectively. Their numerical similarity is defined as:
\begin{equation*}
\text{sim}(\mathbf{x}_i, \mathbf{x}_q) = \frac{\mathbf{v}_i^\top \mathbf{v}_q}{\|\mathbf{v}_i\|_2 \|\mathbf{v}_q\|_2}.
\end{equation*}

To support contrastive analysis, we use label-balanced retrieval under a fixed total budget $K=16$. For classification tasks, we allocate the budget across classes and retrieve the top-$k_c$ most similar instances from each class: $\forall \tilde{y} \in \mathcal{Y}$,
\begin{equation*}
\mathcal{N}_{\tilde{y}}(\mathbf{x}_q) =
\mathop{\mathrm{arg\,top}\text{-}k_{\tilde{y}}}_{\mathbf{x}_i \in \Omega(\mathbf{x}_q),\, y_i = {\tilde{y}}}
\, \text{sim}(\mathbf{x}_i, \mathbf{x}_q).
\end{equation*}
The final retrieved context is the union
$\mathcal{N}(\mathbf{x}_q)=\bigcup_{{\tilde{y}}=0}^{|\mathcal{Y}|-1}\mathcal{N}_{{\tilde{y}}}(\mathbf{x}_q)$,
where $\sum_{{\tilde{y}}=0}^{|\mathcal{Y}|-1} k_{\tilde{y}} = K$.
For regression, we split targets at the global median and apply the same balanced retrieval strategy.

The resulting \emph{Tri-Level Augmented Input} combines dataset statistics, the serialized row, the ground-truth label, and retrieved neighbors. Formally, for each training instance $(\mathbf{x}_i, y_i)$, we define
\begin{equation*}
p_i^* = \mathcal{G}(\mathcal{D}_{\text{train}}) \oplus p_i \oplus y_i \oplus \mathcal{N}(\mathbf{x}_i).
\end{equation*}

\subsection{Tri-Level Rationale Schema}
\label{sec:tri_level_rationale_schema}

To convert the provided evidence into rationales for student fine-tuning, the teacher is prompted to perform in-depth analysis of the Tri-Level Augmented Input $p_i^*$ under a fixed three-part schema: \emph{instance-level reasoning}, \emph{dataset-level reasoning}, and \emph{comparison-level reasoning}. This structure grounds the prediction in the sample itself, contextualizes it with global distributional statistics, and refines it using contrastive evidence from retrieved neighbors. For brevity, we present only the schematic definitions here, while the full prompt template is deferred to Appendix~\ref{sec:appendix_prompt}.

\noindent\textbf{Instance-level reasoning (Self feature-based):}
At this level, the model analyzes each feature strictly using general domain logic and task-relevant common sense. The prompt requires the model to explain its baseline intuition and state the directional impact of each feature on the target variable.

\begin{promptbox}\small
\textbf{(1). Self feature-based reasoning:}
\begin{itemize}[leftmargin=*, nosep]
    \item Analyze each feature from ``Current Sample Features'' strictly using <general domain logic and task-relevant common sense>.
\end{itemize}
\end{promptbox}

\noindent\textbf{Dataset-level reasoning (Global statistics-based):}
At this level, the rationale contextualizes the sample by comparing its feature values with the provided dataset statistics. The prompt requires explicit citation of numerical statistics so that the explanation is grounded in global distributional evidence.

\begin{promptbox}\small
\textbf{(2). Global statistics-based reasoning:}
\begin{itemize}[leftmargin=*, nosep]
    \item Contextualize the sample by comparing its key feature values against the provided dataset statistics for EACH class.
    \item Prioritize those with the clearest class separation or most extreme placement for this sample.
\end{itemize}
\end{promptbox}

\noindent\textbf{Comparison-level reasoning (Similar-cases):}
At this level, the model uses the retrieved neighbors as contrastive anchors to extract feature-combination patterns and critical deviations that further support the prediction. To encourage abstraction rather than record-level reproduction, this part is constrained to two structured blocks: \textit{<Patterns>} and \textit{<Deviations>}. The former summarizes recurring feature-combination alignments, while the latter highlights the most decision-relevant mismatches between the current sample and the retrieved neighbors.
\begin{promptbox}
\small
\textbf{(3). Similar-cases reasoning:}
\begin{itemize}[leftmargin=*, nosep]
    \item Use the provided similar historical cases to extract feature-combination alignment patterns and critical deviations to further strengthen and validate your prediction logic.
\end{itemize}
\end{promptbox}

\subsection{Teacher Generation and Distillation}

\noindent\textbf{Teacher generation.} $\mathcal{M}_T$ denotes a teacher LLM. For each instance $(\mathbf{x}_i, y_i)\in\mathcal{D}_{\text{train}}$, we prompt $\mathcal{M}_T$ with the Tri-Level Augmented Input $p_i^*$ defined in Section~\ref{sec:tri_level_augmented_input} and schema outlined in Section~\ref{sec:tri_level_rationale_schema}. Given the ground-truth label, the teacher acts as an expert that explains \textit{why} the label is plausible, rather than predicting the outcome from scratch. The generation tri-level rationale $r_i$ justifies $y_i$, yielding the fine-tuning corpus $\mathcal{C} = \{(p_i,o_i)\mid (\mathbf{x}_i,y_i)\in\mathcal{D}_{\text{train}}\}$, where $o_i=(r_i,y_i)$.

\noindent\textbf{Student fine-tuning.} The student receives \textit{only} the serialized feature $p_i$ and is trained to generate a unified output sequence formed by concatenating the reasoning trace and label: $[\mathbf{r}_i \oplus y_i]$. We train student parameters $\Theta$ using the standard causal language modeling objective:
\begin{equation*}\resizebox{\hsize}{!}{$
\mathcal{L}_{SFT}(\Theta) = - \frac{1}{|\mathbf{r}_i \oplus y_i|} \sum_{t=1}^{|\mathbf{r}_i \oplus y_i|} \log P_{\Theta}(w_t \mid p_i, w_{<t})$}
\end{equation*}
where $w_t$ represents the $t$-th token in the target sequence. Fine-tuning indirectly injects global distributional knowledge and case-based alignment patterns into the model parameters $\Theta$. During online deployment, the student model operates entirely autonomously without external evidence, making it efficient while preserving reasoning capabilities.

\section{Experiments}
\label{sec:exp}

We answer the following research questions (RQ):

\begin{itemize}[leftmargin=*, nosep]
    \item \textbf{RQ1}: How do base LLMs perform on tabular reasoning, and do they benefit from our proposed evidence sources when provided at inference time?

    \item \textbf{RQ2}: Under raw-feature-only inference, how does TLRD improve base LLMs performance on tabular reasoning, and can it be competitive with strong tabular baselines?
    \item \textbf{RQ3}: Compared with standard LLM distillation, how does TLRD perform in terms of predictive performance and rationale grounding?
    \item \textbf{RQ4}: How do different supervision sources (i.e., self-distillation and teacher-model distillation) affect the performance of TLRD?
    \item \textbf{RQ5}: How small can an LLM be while maintaining competitive performance under TLRD?

\end{itemize}

\subsection{Experiment Setup}

\noindent\textbf{Datasets.} 
We evaluate TLRD on six datasets covering both classification and regression. 
The classification benchmarks are \textsc{Adult} \citep{adult_2}, 
\textsc{Home Credit} (HC) \citep{home-credit-default-risk}, 
\textsc{OkCupid} (OK) \citep{kim2015okcupid}, and 
\textsc{Diabetes130US} (D130) \citep{strack2014impact}, 
spanning social, financial, and clinical domains with varying class imbalance 
and feature heterogeneity. 
For regression, we use \textsc{California} \citep{KELLEYPACE1997291} and 
\textsc{Diamonds} (Dia.) \citep{villanueva2019ggplot2}. 
To reduce computational cost, we downsample \textsc{Diabetes130US} and 
\textsc{Home Credit} to 50k instances each. 
Dataset statistics and preprocessing details are given in Appendix~\ref{sec:appendix_datasets}. 
All datasets are split into training, validation, and test sets with an 8:1:1 ratio.

\noindent\textbf{Models \& Baselines.} 
We consider instruction-tuned LLMs including
\textit{Llama~3.1-8B}~\cite{grattafiori2024llama},
\textit{Gemma~3-12B}~\cite{gemmateam2025gemma3technicalreport},
\textit{Qwen3-8B}, \textit{Qwen3-Next-80B\hspace{0pt}-A3B\hspace{0pt}-Instruct}~\cite{yang2025qwen3},
and \textit{GPT-OSS-20B/120B}~\cite{agarwal2025gpt}. As non-LLM baselines, we use a diverse set of strong tabular methods:
\textit{XGBoost} and \textit{CatBoost} (gradient-boosted decision trees (GBDTs)), \textit{TabM}~\cite{gorishniy2025tabm} (a recent deep tabular model), and
\textit{TabPFN}~\cite{hollmann2023tabpfn} (a tabular foundation model).

\begin{table*}[htb]
\centering
\small
\setlength{\tabcolsep}{2.5pt}

\begin{tabular*}{0.95\textwidth}{@{\extracolsep{\fill}}lrrrrrr@{}}
\toprule
\textbf{Model}
& \textbf{Adult $\uparrow$}
& \textbf{HC $\uparrow$}
& \textbf{D130 $\uparrow$}
& \textbf{OK $\uparrow$}
& \textbf{California $\downarrow$}
& \textbf{Dia. $\downarrow$} \\
\midrule

TabPFN(v2.5) & $68.9$ & $29.7$ & $34.3$ & $55.9$ & $40{,}231.0$ & $\underline{496.0}$ \\
TabPFN(v2.0) & $67.8$ & $24.0$ & $33.3$ & $52.6$ & $\underline{39{,}434.4}$ & $505.7$ \\
XGBoost      & $\underline{72.0}$ & $\underline{29.8}$ & $42.7$ & $58.8$ & $43{,}253.8$ & $527.7$ \\
CatBoost     & $71.3$ & $28.1$ & $42.8$ & $\underline{58.9}$ & $42{,}654.1$ & $502.6$ \\
TabM         & $70.1$ & $28.0$ & $\underline{43.7}$ & $58.2$ & $48{,}855.8$ & $519.7$ \\

\midrule

\textbf{GPT-OSS-20B (Zero-shot)}
& $57.9$ & $16.6$ & $14.2$ & $49.4$ & $201{,}829.8$ & $2{,}477.1$ \\
\hspace{1em}Tri-level Reasoning
& $\mathbf{66.9}$ & $24.3$ & $36.0$ & $54.5$ & $75{,}992.7$ & $705.1$ \\

\textbf{Qwen3-Next-80B (Zero-shot)}
& $61.9$ & $15.5$ & $8.9$ & $44.5$ & $333{,}534.9$ & $1{,}289.5$ \\
\hspace{1em}Tri-level Reasoning
& $63.5$ & $23.5$ & $30.4$ & $52.6$ & $73{,}383.1$ & $698.4$ \\

\textbf{GPT-OSS-120B (Zero-shot)}
& $58.2$ & $16.2$ & $21.1$ & $50.4$ & $114{,}749.6$ & $2{,}632.1$ \\
\hspace{1em}Tri-level Reasoning
& $66.6$ & $25.2$ & $37.3$ & $55.0$ & $\mathbf{62{,}906.6}$ & $\mathbf{649.7}$ \\

\midrule

\textbf{Llama 3.1-8B (Zero-shot)}
& $51.3$ & $15.5$ & $19.7$ & $46.0$ & $3.02 _{\times 10^{7}}$ & $3.64 _ {\times 10^{8}}$ \\
\hspace{1em}Tri-level Reasoning
& $63.9$ & $17.8$ & $27.1$ & $51.5$ & $1.12 _{\times 10^{7}}$ & $2{,}193.4$ \\
\hspace{1em}TLRD (GPT-OSS-120B)
& $64.7$ & $25.1$ & $34.9$ & $54.2$ & $86{,}572.7$ & $1{,}531.9$ \\
\hspace{1em}TLRD (Qwen3-Next-80B)
& $63.1$ & $22.7$ & $35.7$ & $53.2$ & $112{,}008.0$ & $1{,}453.7$ \\

\textbf{Qwen3-8B (Zero-shot)}
& $57.9$ & $20.4$ & $6.1$ & $53.6$ & $175{,}459.5$ & $2{,}280.9$ \\
\hspace{1em}Tri-level Reasoning
& $65.5$ & $26.0$ & $36.6$ & $54.0$ & $65{,}660.6$ & $691.0$ \\
\hspace{1em}TLRD (GPT-OSS-120B)
& $65.7$ & $\mathbf{26.3}$ & $33.4$ & $55.6$ & $86{,}341.3$ & $1{,}316.6$ \\
\hspace{1em}TLRD (Qwen3-Next-80B)
& $63.0$ & $22.7$ & $36.9$ & $55.0$ & $91{,}877.5$ & $1{,}326.1$ \\

\textbf{Gemma 3-12B (Zero-shot)}
& $55.1$ & $15.0$ & $9.2$ & $50.3$ & $568{,}304.3$ & $2{,}589.5$ \\
\hspace{1em}Tri-level Reasoning
& $63.1$ & $18.8$ & $33.2$ & $51.1$ & $70{,}633.5$ & $1{,}044.9$ \\
\hspace{1em}TLRD (GPT-OSS-120B)
& $63.4$ & $25.7$ & $34.5$ & $\mathbf{56.1}$ & $85{,}646.6$ & $945.9$ \\
\hspace{1em}TLRD (Qwen3-Next-80B)
& $64.8$ & $22.3$ & $\mathbf{38.1}$ & $55.6$ & $96{,}473.5$ & $1{,}060.4$ \\

\bottomrule
\end{tabular*}
\caption{Results across all datasets. Bold values denote the best result among LLM-based methods, and underlined values denote the strongest non-LLM tabular baseline for each dataset. Among fine-tuned student backbones, TLRD variants achieve the strongest classification performance within every backbone group.}
\vspace{-0.14cm}
\label{tab:combined_by_backbone}
\end{table*}

\noindent\textbf{Evaluation Metrics.} For binary classification tasks, given the prevalence of class imbalance in financial and risk datasets (e.g., \textsc{Home Credit}), we report \textit{F1-score}. For multi-class classification, we report \textit{Macro-F1}, i.e., the unweighted average of per-class F1 scores. For regression tasks, we consider \textit{Root Mean Squared Error (RMSE)}.

\noindent\textbf{Implementation Details.} 
Experiments are run on \mbox{4$\times$A100} GPUs.
For \textit{XGBoost}, \textit{CatBoost}, and \textit{TabM} baselines, we perform hyperparameter optimization on the validation set with 50 Optuna trials; the search spaces are in Appendix Table~\ref{tab:hparam-spaces-upd}.
Due to limited computation resources, we conduct LoRA fine-tuning~\citep{hu2022lora} on three relatively compact models: \textit{Llama~3.1-8B}, \textit{Qwen3-8B}, and \textit{Gemma~3-12B}. Full details are provided in Appendix~\ref{sec:appendix_expsetup}.

\subsection{Results}
\noindent\textbf{RQ1: Baseline LLM performance and impact of extra evidence sources.}
In base LLMs, we follow \textit{TabLLM}~\cite{xie2024finben, pmlr-v206-hegselmann23a, wolff2025well} and use the \textbf{zero-shot} setting as baseline. We then study a \textbf{Tri-level Reasoning} variant, in which the query is augmented with dataset statistics and retrieved similar samples to provide tri-level context at inference time.

\noindent\textbf{Baseline performance.}
As shown in Table \ref{tab:combined_by_backbone}, base LLMs exhibit weak and unstable zero-shot performance. They perform better on \textsc{Adult} and \textsc{OK}, where patterns are usually more aligned with common social semantics that LLMs have captured during pre-training. In contrast, performance drops substantially on \textsc{HC} and \textsc{D130}, which rely more on dataset-specific statistical regularities, class imbalance, and specialized feature semantics. For example, \textit{Qwen3-Next-80B} achieves only 8.9 on \textsc{D130}. The limitation is even more pronounced on regression, where several models produce extremely large RMSE values, such as \textit{Llama~3.1-8B}, which reaches $3.02 \times 10^{7}$ on \textsc{California}. Tabular regression requires accurate scale awareness and fine-grained numerical mappings from inputs to targets. Without explicit grounding, LLMs can easily produce severe scale mismatches.

\noindent\textbf{Tri-level Reasoning performance.}
Tri-level Reasoning improves the performance of base LLMs across all models and datasets. For example, \textit{GPT-OSS-20B} improves from 14.2 to 36.0 on \textsc{D130}. For the challenges faced by zero-shot, dataset-level reasoning provides dataset statistical context, allowing the model to \textbf{judge which class a feature value is more aligned with and how informative it is.} Comparison-level reasoning further adds local contrastive evidence through balanced retrieved neighbors, helping the model \textbf{capture discriminative feature combinations}. These benefits also extend to regression. Dataset-level reasoning provides coarse calibration signals, while comparison-level reasoning supplies exact target-value references. Together, they \textbf{help the model better capture the output scale and infer a more plausible prediction range.} This motivates our next hypothesis: can we let a strong teacher analyze these evidence sources more deeply under the ground-truth label, and then distill the resulting evidence-grounded knowledge into a compact student model?

\paragraph{RQ2: TLRD performance.} Motivated by this hypothesis, TLRD uses the tri-level schema during corpus construction to let a teacher model generate evidence-grounded analyses conditioned on the given label, and distills the resulting knowledge into the student through fine-tuning. Although the fine-tuned student receives only raw features, \textbf{it consistently achieves the strongest classification performance within its backbone family.} For example, under \textit{GPT-OSS-120B} supervision, \textit{Qwen3-8B} reaches 26.3 on \textsc{HC}. Under \textit{Qwen3-Next-80B} supervision, \textit{Gemma~3-12B} reaches 38.1 on \textsc{D130}. Compared with zero-shot, these gains are substantial, showing that TLRD can effectively convert teacher-side evidence analysis into useful supervision for compact student models. On regression, TLRD still improves substantially over zero-shot, but it often remains below Tri-level Reasoning. Regression requires much finer scale control and numerical calibration than discrete class prediction. Compressing such precise quantitative mappings into the student parameters remains challenging.

Table~\ref{tab:combined_by_backbone} shows that \textbf{TLRD substantially narrows the gap between LLM-based methods and strong tabular baselines on classification tasks.} On \textsc{OK}, TLRD achieves a score of 56.1, slightly surpassing \textit{TabPFN} v2.5 (55.9) and approaching \textit{TabM} (58.2), \textit{CatBoost} (58.9), and \textit{XGBoost} (58.8). On \textsc{D130}, all three fine-tuned backbones outperform both \textit{TabPFN} variants, further showing that TLRD is competitive with modern tabular models.

\begin{figure*}[!t]
    \centering
    \includegraphics[width=\textwidth]{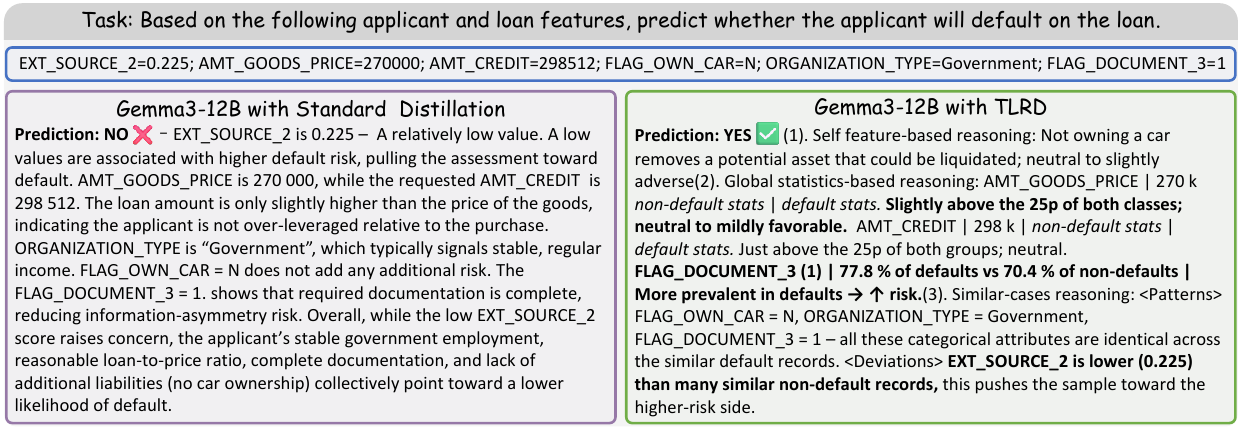}
    \caption{A qualitative case study comparing the same model under standard distillation and TLRD. For a sample with ground-truth label \texttt{YES}, the standard distillation model predicts incorrectly, whereas TLRD predicts correctly and provides a more grounded tri-level rationale.}
    \vspace{-0.1cm}
    \label{fig:rq3_compare_gemma}
\end{figure*}

\paragraph{RQ3: Comparison between TLRD and standard rationale distillation.}
For this comparison, we fix \textit{GPT-OSS-120B} as the teacher model for corpus construction. The standard distillation is trained on teacher-generated free-form explanations.

\noindent\textbf{Predictive performance.}
Table~\ref{tab:rq3_standard_vs_tlrd} shows that TLRD outperforms standard distillation across all three backbones on the classification datasets. On \textsc{Dia.}, the advantage is less uniform, but TLRD still achieves the strongest overall result. These findings suggest that TLRD provides more effective supervision by organizing the explanation around tri-level evidence under the designed schema.

\begin{table}[!htbp]
\centering
\scriptsize
\setlength{\tabcolsep}{2pt}
\renewcommand{\arraystretch}{0.8}
\begin{tabular*}{\columnwidth}{@{\extracolsep{\fill}}lccc}
\toprule
Method & HC $\uparrow$ & OK $\uparrow$ & Dia. $\downarrow$ \\
\midrule
Llama~3.1-8B Standard Distill. & 19.9 & 46.8 & 1201.7 \\
Llama~3.1-8B TLRD              & 25.1 & 54.2 & 1531.9 \\
Qwen3-8B Standard Distill.     & 23.7 & 51.9 & 1317.5 \\
Qwen3-8B TLRD                  & 26.3 & 55.6 & 1316.6 \\
Gemma~3-12B Standard Distill.  & 20.0 & 44.8 & 1152.1 \\
Gemma~3-12B TLRD               & 25.7 & 56.1 & 945.9 \\
\bottomrule
\end{tabular*}
\caption{Performance of standard distillation and TLRD.}

\vspace{-0.1cm}
\label{tab:rq3_standard_vs_tlrd}
\end{table}

\noindent\textbf{Rationale grounding.}
Figure~\ref{fig:rq3_compare_gemma} compares two \textit{Gemma~3-12B} variants on the same \textsc{HC} instance. Although standard distillation generates a fluent explanation, its reasoning is largely based on generic commonsense assumptions that appear plausible but are not supported by concrete evidence. By contrast, TLRD structures its explanation into instance-level, dataset-level, and comparison-level reasoning, yielding a more grounded explanation. At the dataset level, it captures the counterintuitive but dataset-supported pattern that \texttt{FLAG\_DOCUMENT\_3 = 1} is more prevalent among default applicants. At the comparison level, identifies \texttt{EXT\_SOURCE\_2 = 0.225} as a key deviation from similar non-default profiles. Together, these elements indicate that TLRD combines reasonable local intuition with dataset-specific evidence, rather than relying only on potentially misleading narrative associations.

\begin{table}[!t]
\centering
\scriptsize
\renewcommand{\arraystretch}{0.8}
\setlength{\tabcolsep}{1mm} 
\begin{tabular*}{\columnwidth}{@{\extracolsep{\fill}}llccc@{}}
\toprule
Student & Teacher & HC $\uparrow$ & OK $\uparrow$ & Dia. $\downarrow$ \\
\midrule
\textbf{Llama 3.1-8B} & Self           & 9.7  & 52.6 & 1606.1 \\
                       & GPT-OSS-20B    & 24.3 & 54.8 & 1475.3 \\
                       & Qwen3-Next-80B & 22.7 & 53.2 & 1453.7 \\
                       & GPT-OSS-120B   & 25.1 & 54.2 & 1531.9 \\
                       & GPT-5.2        & 25.0 & 55.6 & 1391.7 \\
\midrule
\textbf{Qwen3-8B}      & Self           & 24.9 & 53.2 & 1417.2 \\
                       & GPT-OSS-20B    & 25.3 & 56.3 & 1135.7 \\
                       & Qwen3-Next-80B & 22.7 & 55.0 & 1326.1 \\
                       & GPT-OSS-120B   & 26.3 & 55.6 & 1316.6 \\
                       & GPT-5.2        & 24.7 & 54.4 & 1295.5 \\
\midrule
\textbf{Gemma 3-12B}   & Self           & 20.8 & 56.0 & 990.1 \\
                       & GPT-OSS-20B    & 24.8 & 53.5 & 859.6 \\
                       & Qwen3-Next-80B & 22.3 & 55.6 & 1060.4 \\
                       & GPT-OSS-120B   & 25.7 & 56.1 & 945.9 \\
                       & GPT-5.2        & 24.4 & 57.0 & 906.2 \\
\bottomrule
\end{tabular*}
\caption{Performance under different supervision sources for each student backbone.}

\vspace{-0.1cm}
\label{tab:rq5_teacher_comparison}
\end{table}

\paragraph{RQ4: Effect of different supervision sources.} We consider self-distillation and distillation from multiple teacher models, including \textit{GPT-OSS-20B}, \textit{Qwen3-Next-80B}, \textit{GPT-OSS-120B}, and \textit{GPT-5.2}.

Table~\ref{tab:rq5_teacher_comparison} shows that under the TLRD framework, mid-sized teacher \textit{GPT-OSS-20B} already provides strong supervision, and self-distillation can also be competitive in some cases. Meanwhile, performance does not always scale with the supervisors, meaning that TLRD typically does not require very large teacher models. In several settings, relatively smaller supervisors even achieve performance similar to larger ones. Overall, for tabular tasks, \textbf{TLRD does not necessarily require massive teacher LLMs. Rather, it can successfully rely on a mid-sized teacher model to enable efficient and resource-friendly deployment.}

\begin{table}[!h]
\centering
\tiny
\setlength{\tabcolsep}{1pt}
\renewcommand{\arraystretch}{0.78}
\begin{tabular*}{\columnwidth}{@{\extracolsep{\fill}}lccccc|ccccc@{}}
\toprule
& \multicolumn{5}{c|}{Gemma 3} 
& \multicolumn{5}{c}{Qwen3} \\
\cmidrule(lr){2-6} \cmidrule(lr){7-11}
Dataset 
& 27B & 12B & 4B & 1B & 270M 
& 14B & 8B & 4B & 1.7B & 0.6B \\
\midrule
HC $\uparrow$ 
& 26.6 & 25.7 & 23.9 & 13.7 & 9.5 
& 26.1 & 26.3 & 24.2 & 23.4 & 23.1 \\

OK $\uparrow$ 
& 59.7 & 57.0 & 56.5 & 40.9 & 34.1 
& 56.8 & 56.3 & 55.7 & 55.4 & 43.4 \\

Dia. $\downarrow$ 
& 712.8 & 859.6 & 1376.9 & 4457.2 & 14932.6 
& 879.2 & 1135.7 & 1507.0 & 2008.7 & 3421.9 \\
\bottomrule
\end{tabular*}
\caption{Performance of \textit{Qwen3} and \textit{Gemma 3} families. The selected supervisors are \textit{GPT-OSS-120B} for \textsc{HC}, \textit{GPT-5.2} for \textit{Gemma 3} on \textsc{OK}, and \textit{GPT-OSS-20B} for \textit{Qwen3} on \textsc{OK} and for both families on \textsc{Dia}.}
\label{tab:scale_sensitivity}
\vspace{-1mm}
\end{table}

\paragraph{RQ5: Performance of models from the same family at different scales.}  We conduct experiments on the \textit{Qwen3} and \textit{Gemma 3} families. Since our focus is on how small the student can be while maintaining performance, we report the results obtained with the best teacher supervisor.

Table~\ref{tab:scale_sensitivity} summarizes the results across three representative datasets. Overall, increasing the student model scale generally improves TLRD performance. In addition, both 4B-scale students still show strong performance on \textsc{OK} and only mild degradation on \textsc{HC}. A particularly encouraging result appears on \textsc{OK}, where \textit{Gemma 3-27B} reaches 59.7 Macro-F1, \textbf{even surpassing the strongest baseline, \textit{CatBoost} (58.9)}.

Combined with the analysis in RQ4, these results show that TLRD offers a potential practical direction for resource-constrained scenarios: \textbf{using a 4B-scale student model supervised by a mid-sized teacher to enable effective and resource-friendly deployment under classification tasks}, while larger students or stronger teachers can be used when further performance gains are needed.

\subsection{{Design and Analyses on Hallucination, Consistency, and Human Annotation; Ablation Study; Further Discussion}}
\paragraph{Design and Analyses.}
We present design safeguards and conduct additional quantitative analyses in Appendix~\ref{analyse of usefulness}, including hallucination, consistency, and human annotation, to further assess the reliability and usefulness of TLRD. 
\paragraph{Ablation.} 
We provide ablations in Appendix~\ref{sec:appendix_ablation} to examine key designs of TLRD, including extra evidence sources and the designed schema.
\paragraph{Further Discussion.}
From a performance perspective, both tree-based methods and TabM have mature optimization pipelines and class-imbalance handling, whereas we did not conduct systematic hyperparameter optimization for LLMs due to the high cost and time-consuming. However, TLRD still surpasses them with \textit{Gemma 3-27B} on \textsc{OK}, which demonstrates the potential of TLRD. We also explored several task-specific optimization directions in Appendix~\ref{sec:appendix_task_specific_optimization}. We hope our work encourages more mature and practical deployment of LLM-based methods for tabular tasks.

\section{Conclusion}
We proposed TLRD, a framework that converts label-only tabular datasets into structured rationale supervision for LLMs. TLRD provides the teacher with a Tri-Level Augmented Input, which extends the labeled sample with dataset statistics and retrieved neighbors, and uses a Tri-Level Schema to guide grounded tri-level rationale generation. By distilling these rationales into student models, TLRD enables efficient inference from raw features only. Experiments show that TLRD achieves competitive performance against strong tabular baselines, while producing grounded and human-readable explanations. Overall, TLRD provides a practical assistive tool for supporting transparent decision support, highlighting structured rationale distillation as an effective bridge between tabular prediction and explanation.

\section*{Limitations}
Our study has several limitations that suggest caution and opportunities for future work.
\begin{itemize}[leftmargin=*, nosep]

\item \textbf{Distillation of Spurious Correlations.} TLRD depends on the quality of teacher-generated rationales. Although we condition the teacher on the ground-truth label to anchor the rationale to the correct outcome, the teacher may still produce spurious correlations, subtle logical leaps, or plausible but unsupported justifications. Since the student is trained to mimic this corpus, undetected artifacts in teacher outputs may be distilled into the student model.

\item \textbf{Scalability to High-Dimensional Features and Classes.} For datasets with many columns, we rely on feature selection to construct concise natural-language prompts, which may omit long-tail variables or weak feature interactions. In addition, our dataset-level reasoning schema appends class-specific statistics; when the number of classes is large, this can substantially increase input length and dilute the useful context. Nevertheless, many practical tabular decision tasks are binary or low-class classification problems, where TLRD remains directly applicable.

\item \textbf{Use of LoRA Instead of Full Supervised Fine-Tuning.}
Due to limited computational resources, we adopt LoRA-based parameter-efficient fine-tuning rather than full supervised fine-tuning (full SFT). While this choice enables training on multiple datasets and backbones within a practical budget, it also limits our ability to examine whether full SFT could yield larger improvements in predictive performance or rationale quality. Since full-parameter adaptation may internalize distilled knowledge differently, this remains an important direction for future work.
\item \textbf{Inherent Limitations of LLM-Generated Rationales and Human Evaluation.}
TLRD inherits the general limitations of LLM-based generation, including possible hallucination, over-confident wording, and sensitivity to prompt or decoding settings. We conduct quantitative analyses and a small-scale human study to assess whether the generated rationales are grounded, consistent, and useful. However, human evaluation of tabular rationales is inherently task- and domain-dependent, and a larger study with domain experts would be needed to fully assess their impact in real-world decision-making. Therefore, TLRD should be deployed strictly as a decision-support tool, not as an autonomous decision-maker.

\end{itemize}

\section*{Ethical Considerations}

\begin{itemize}
\item \textbf{Bias Inheritance.} Tabular datasets frequently include sensitive attributes like race, gender, and socioeconomic status. There is a risk that the teacher generates biased outcomes and leads the student model to learn biased rationales. Practitioners may conduct rigorous algorithmic fairness audits on the data before applying the TLRD pipeline.

\item \textbf{Data Privacy.} Our methodology makes a concerted effort to protect data privacy. By explicitly prohibiting the teacher model from referencing neighbor IDs or precise historical records, and by completely discarding the retrieved neighbors during the student fine-tuning phase, TLRD structurally mitigates the risk of verbatim training data leakage at deployment. Nevertheless, there remains a theoretical risk that highly unique feature combinations (outliers) could be implicitly memorized and reconstructed in the generalized \textit{<Patterns>} or \textit{<Deviations>} blocks.

\end{itemize}
\bibliography{custom}

\clearpage
\appendix
\section{Datasets}
\label{sec:appendix_datasets}

\begin{table*}[!t]
\centering
\small
\setlength{\tabcolsep}{4pt}
\begin{tabular}{l l r r r c l l}
\toprule
\textbf{Dataset} & \textbf{Task} & \textbf{$N$ (raw)} & \textbf{\#Num} & \textbf{\#Cat} & \textbf{\#Classes} & \textbf{Imbalance (\%)} & \textbf{License} \\
\midrule
Adult & Binary cls. & 48{,}842 & 6 & 8 & 2 & 23.9:76.1 & CC BY 4.0 \\
Home Credit & Binary cls. & 307{,}511 & 106 & 16 & 2 & 8.1:91.9 & Competition data \\
OkCupid & Multi cls. & 50{,}789 & 2 & 17 & 3 & 71.6:18.8:9.6 & Public-permissioned \\
Diabetes130US & Multi cls. & 69{,}990 & 8 & 37 & 3 & 59.3:31.7:9.0 & CC BY 4.0 \\
\midrule
California & Regression & 20{,}640 & 8 & 1 & -- & -- & CC0: Public Domain \\
Diamonds & Regression & 53{,}940 & 6 & 3 & -- & -- & CC BY 4.0 \\
\bottomrule
\end{tabular}
\caption{Dataset statistics. For binary datasets, \textbf{Imbalance (\%)} reports the positive-to-negative class ratio (P:N) in percentages. Preprocessing and subsampling details are provided in Appendix~\ref{sec:appendix_datasets}.}
\label{tab:dataset_stats_raw}

\end{table*}

We use six real-world tabular datasets: two binary classification, two multi-class classification, and two regression datasets. Below we provide detailed dataset descriptions and preprocessing details. Unless stated otherwise, we use the full dataset and partition it into training, validation, and test splits with an 8:1:1 ratio. Summary statistics are reported in Table~\ref{tab:dataset_stats_raw}. 

\noindent\textbf{Adult \citep{adult_2}.}\footnote{\url{https://archive.ics.uci.edu/dataset/2/adult}}
A binary classification dataset from the 1994 U.S.\ Census, where the goal is to predict whether an individual's annual income exceeds \$50K based on demographic and employment attributes.

\noindent\textbf{Home Credit \citep{home-credit-default-risk}.}\footnote{\url{https://kaggle.com/competitions/home-credit-default-risk}}
A binary credit-risk dataset where the goal is to predict whether an applicant defaults on the loan. The dataset contains 307{,}511 labeled applications, is highly imbalanced, with only about 8.07\% positive (default) cases, and includes 122 features. During LLM inference and corpus construction, we retain only the top 15 features ranked by CatBoost SHAP importance. To mitigate optimization bias from severe class imbalance during LLM fine-tuning while also satisfying the computational constraints of Transformer-based in-context inference with long contexts, we apply undersampling on the training split when building LLM inference corpora and when forming TabPFN \citep{hollmann2023tabpfn} contexts. Specifically, we perform ratio-preserving undersampling with a $3{:}1$ negative-to-positive target and continue sampling until the training subset reaches 40{,}000 examples; this reduces majority-class dominance while keeping a reasonable approximation to the original class prior, as well as token budget, GPU memory usage, and training time tractable. This treatment is particularly important for TabPFN: beyond the quadratic context-length cost of concatenating labeled training examples in a single forward pass, using the original highly imbalanced pool without undersampling makes in-context predictions strongly majority-class biased, leading to systematic under-detection of defaults. For evaluation efficiency, we further subsample the test split to 5{,}000 examples while preserving the original class ratio, improving throughput and reproducibility without altering the underlying data distribution.

\noindent\textbf{OkCupid \citep{kim2015okcupid}.}\footnote{\url{https://www.openml.org/d/42734}}
A multi-class classification dataset where the goal is to predict the profile author's occupation category (\textsc{stem} / non-\textsc{stem} / student). Inputs are mixed-type attributes describing demographics and self-reported profile fields, with a predominance of categorical variables.

\noindent\textbf{Diabetes130US (1999--2008) \citep{strack2014impact}.}\footnote{\url{https://archive.ics.uci.edu/dataset/296/diabetes+130-us+hospitals+for+years+1999-2008}}
A multi-class clinical prediction dataset containing inpatient encounters for diabetes patients from 130 U.S.\ hospitals (1999--2008), with 101{,}766 instances. We consider the original readmission prediction setting, where the label indicates the readmission outcome as multiple categories (commonly \texttt{NO}, \texttt{<30}, and \texttt{>30}). Data preprocessing follows the cohort-construction logic of Strack et al. Specifically, we remove encounters with hospice/death discharge outcomes, keep only the first encounter per patient to ensure patient-level independence, drop high-missing columns (\texttt{weight} and \texttt{payer\_code}), and impute missing \texttt{medical\_specialty} as \texttt{missing}. We then follow Strack-style binning to map ICD-9 diagnosis codes into coarse clinical groups and to rebin age into three categories. Our downstream setup intentionally differs from Strack et al. Their main aim is explanatory statistical modeling (logistic-regression-based analysis of HbA1c and readmission), whereas our main aim is predictive classification. We use the original 3-class prediction target, \texttt{readmitted} $\in \{\texttt{<30}, \texttt{>30}, \texttt{NO}\}$. The dataset includes mixed numerical and categorical attributes describing demographics, diagnoses, procedures, lab tests, and medications. After preprocessing, we retain 69{,}990 samples with 45 features. The full preprocessed dataset is used for training and evaluating the non-LLM baselines. During LLM inference and corpus construction, we retain only the top 15 features ranked by CatBoost SHAP importance. Using an 8:1:1 split, we further subsample the training set to 40{,}000 examples for corpus construction and the test set to 5{,}000 examples for efficient LLM inference.

\noindent\textbf{California \citep{KELLEYPACE1997291}.}\footnote{\url{https://www.kaggle.com/datasets/camnugent/california-housing-prices}}
A regression dataset derived from the 1990 California census, where the target is the median house value for a census block group. We use the Kaggle version of the dataset released as \texttt{camnugent/california-housing-prices}.

\noindent\textbf{Diamonds \citep{villanueva2019ggplot2}.}\footnote{\url{https://doi.org/10.5281/zenodo.3522106}}
A regression dataset of round-cut diamonds with features describing the ``4 Cs'' (carat, cut, color, clarity) and physical measurements (e.g., depth, table, and x/y/z), and the target being price in USD. We follow standard setup steps consistent with common tabular benchmarks.

\section{Experimental Setup}
\label{sec:appendix_expsetup}

\subsection{Baselines}
\label{sec:appendix_baselines}

We use a diverse set of strong tabular baselines:
XGBoost~\citep{chen2016xgboost} and CatBoost~\citep{prokhorenkova2018catboost} (gradient-boosted decision trees (GBDTs)), TabM~\citep{gorishniy2025tabm} (deep tabular model), and TabPFN~\citep{hollmann2023tabpfn} (foundation tabular predictor).

\noindent\textbf{TabPFN (v2.5 / v2.0).}
TabPFN is a transformer-based foundation model for tabular prediction. It is pretrained on a large distribution of synthetic supervised learning tasks and performs inference in an in-context manner, conditioning directly on the observed training data to make predictions for test instances. We use the latest major release (TabPFN-2.5) and TabPFN-v2 (v2.0). Both are evaluated using their standard inference protocol.

\noindent\textbf{XGBoost and CatBoost.}
XGBoost and CatBoost are gradient boosting decision tree methods that construct an ensemble of trees sequentially, with each new tree trained to correct the errors of the current ensemble; CatBoost additionally includes specialized handling of categorical features. For both XGBoost and CatBoost, we reference the hyperparameter spaces reported in TabArena \citep{erickson2026tabarena}. We perform HPO using the TPE sampler in Optuna \citep{akiba2019optuna} with 50 trials for each dataset--model pair, select the best configuration on the validation split, and then evaluate that selected configuration on the test split. For XGBoost and CatBoost, we use early stopping with a patience of 300 rounds. Hyperparameter details are in Table~\ref{tab:hparam-spaces-upd}. To address class imbalance, we apply fixed loss weighting; implementation details are provided in Table~\ref{tab:imbalance-handling}.

\begin{table*}[!t]
\centering
\small
\setlength{\tabcolsep}{5pt}
\begin{tabular}{l l l}
\toprule
\textbf{Model} & \textbf{Hyperparameter} & \textbf{Search Space / Setting} \\
\midrule
\multirow{12}{*}{XGBoost} & learning\_rate & LogUniform([0.005, 0.1]) \\
 & max\_depth & LogUniformInt([4, 10]) \\
 & min\_child\_weight & LogUniform([0.001, 5.0]) \\
 & subsample & Uniform([0.6, 1.0]) \\
 & colsample\_bylevel & Uniform([0.6, 1.0]) \\
 & colsample\_bynode & Uniform([0.6, 1.0]) \\
 & reg\_alpha & Uniform([1e-4, 5.0]) \\
 & reg\_lambda & Uniform([1e-4, 5.0]) \\
 & grow\_policy & Choice([\texttt{depthwise}, \texttt{lossguide}]) \\
 & max\_cat\_to\_onehot & LogUniformInt([8, 100]) \\
 & max\_leaves & LogUniformInt([8, 1024]) \\
 & n\_estimators & Fixed to 10{,}000 \\
\midrule
\multirow{14}{*}{CatBoost} & learning\_rate & LogUniform([0.005, 0.1]) \\
 & bootstrap\_type & Bernoulli \\
 & subsample & Uniform([0.7, 1.0]) \\
 & grow\_policy & Choice([\texttt{SymmetricTree}, \texttt{Depthwise}]) \\
 & depth & UniformInt([4, 8]) \\
 & colsample\_bylevel & Uniform([0.85, 1.0]) \\
 & l2\_leaf\_reg & LogUniform([1e-4, 5]) \\
 & leaf\_estimation\_iterations & LogUniformInt([1, 20]) \\
 & one\_hot\_max\_size & LogUniformInt([8, 100]) \\
 & model\_size\_reg & LogUniform([0.1, 1.5]) \\
 & max\_ctr\_complexity & UniformInt([2, 5]) \\
 & boosting\_type & Plain \\
 & max\_bin & 254 \\
 & n\_estimators & Fixed to 10{,}000 \\
\midrule
\multirow{8}{*}{TabM} & embedding\_method & Fixed to \texttt{piecewise linear embedding} \\
 & k & Const([32]) (not tuned) \\
 & n\_blocks & UniformInt([1, 4]) \\
 & d\_block & UniformInt([64, 1024], step=16) \\
 & lr & LogUniform([1e-4, 5e-3]) \\
 & weight\_decay & $\{0, \text{LogUniform}[1\mathrm{e}{-4}, 1\mathrm{e}{-1}]\}$ \\
 & n\_bins & UniformInt([2, 128]) \\
 & d\_embedding & UniformInt([8, 32], step=4) \\
\bottomrule
\end{tabular}
\caption{Hyperparameter search spaces used in HPO.}
\label{tab:hparam-spaces-upd}
\end{table*}

\begin{table*}[!t]
\centering
\small
\setlength{\tabcolsep}{6pt}
\begin{tabular}{l l l}
\toprule
\textbf{Model} & \textbf{Task} & \textbf{Balancing Rule} \\
\midrule
\multirow{2}{*}{XGBoost} & Binary & \texttt{scale\_pos\_weight = N\_neg / N\_pos} \\
 & Multiclass & \texttt{w\_k = N\_max / N\_k}, \texttt{sample\_weight\_i = w\_{y\_i}} \\
\midrule
\multirow{2}{*}{CatBoost} & Binary & \texttt{class\_weights = [1.0, N\_neg / N\_pos]} \\
 & Multiclass & \texttt{class\_weights[k] = N\_max / N\_k} \\
\midrule
\multirow{2}{*}{TabM} & Binary & \texttt{pos\_weight = N\_neg / N\_pos} \\
 & Multiclass & \texttt{class\_weights[k] = N\_max / N\_k} \\
\bottomrule
\end{tabular}
\caption{Class imbalance handling rules used for baselines (fixed, not tuned).}
\label{tab:imbalance-handling}
\end{table*}
\noindent\textbf{TabM.}
TabM is a parameter-efficient ensemble-based tabular deep learning model. It efficiently emulates an ensemble of MLPs by training multiple predictors in parallel while sharing a large portion of their parameters. Hyperparameter settings are reported in Table~\ref{tab:hparam-spaces-upd}, and class-imbalance handling is reported in Table~\ref{tab:imbalance-handling}. Our configuration follows the official TabM repository recommendations. We used early stopping on the validation set with a patience of 16 epochs, selecting the checkpoint with the best validation selection score; validation was evaluated after every epoch, and training stopped if the score did not improve for 16 consecutive epochs.

\subsection{LLM Hyperparameters}
\label{sec:appendix_llm_hyperparameters}

\noindent\textbf{Base LLM backbones.}
We use the same configuration for both base-model inference and corpus construction. We set the maximum context length to 32{,}768 tokens, use temperature $0.6$, and apply nucleus sampling with $p=0.95$.

\noindent\textbf{LLM Fine-tuning.}
We use parameter-efficient fine-tuning (PEFT) with LoRA~\citep{hu2022lora}, implemented with LLaMA-Factory~\citep{zheng2024llamafactory}. Fine-tuning hyperparameters are summarized in Table~\ref{tab:llm_finetune_hparams}.

\begin{table*}[!t]
\centering
\small
\setlength{\tabcolsep}{6pt}
\begin{tabular}{l l}
\toprule
\textbf{Hyperparameter} & \textbf{Value} \\
\midrule
learning\_rate & 0.0001 \\
train\_batch\_size & 4 \\
eval\_batch\_size & 8 \\
seed & 42 \\
distributed\_type & multi-GPU \\
num\_devices & 4 \\
gradient\_accumulation\_steps & 8 \\
lr\_scheduler\_type & cosine \\
num\_epochs & 1.0 (only California: 2.0) \\
optimizer & \texttt{adamw\_torch} \\
lora\_rank & 8 \\
lora\_alpha & 16 \\
lora\_dropout & 0 \\
lora\_target & \texttt{all} \\
\bottomrule
\end{tabular}
\caption{LLM fine-tuning hyperparameters.}
\label{tab:llm_finetune_hparams}
\end{table*}

\noindent\textbf{Fine-tuned LLM inference.}
After dataset-specific fine-tuning, the model is already aligned to the target prediction task. Therefore, we change the temperature to $0$ to obtain stable and deterministic predictions.

\section{Prompt}
\label{sec:appendix_prompt}
We use different prompt templates for (i) zero-shot inference, (ii) w/o Statistics, (iii) w/o Comparison, (iv) Tri-level Reasoning, (v) standard fine-tuning (FT) corpus construction, (vi) TLRD corpus construction, and (vii) TLRD model inference. The zero-shot template is also used for inference with the standard fine-tuned model. After each template, we list feature explanations and feature importance (high to low), ordered by CatBoost SHAP importance. Bold angle-bracketed tokens (e.g., \textbf{\texttt{<DATASET\_DOMAIN>}}) are instantiated with dataset-specific metadata/domain knowledge. Specifically, templates for (i)--(iv) are shown in Table~\ref{tab:prompt-zero-abl-tlr}, template (v) is shown in Table~\ref{tab:prompt-ft-corpus}, template (vi) is shown in Table~\ref{tab:prompt-tlrd-corpus}, and template (vii) is shown in Table~\ref{tab:prompt-tlrd-infer}.

\lstset{
  basicstyle=\ttfamily\small,
  columns=fullflexible,
  breaklines=true,
  breakatwhitespace=true,
  keepspaces=true,
  upquote=true,
  showstringspaces=false
}

\newtcolorbox{promptcard}[1]{%
  enhanced,
  colframe=gray!40,
  colback=gray!10,
  arc=2pt,
  boxrule=0.3pt,
  left=4pt,right=4pt,top=4pt,bottom=4pt,
  colbacktitle=gray!20,
  coltitle=black,
  fonttitle=\bfseries,
  title=#1,
  toptitle=2mm, bottomtitle=2mm
}

\begin{table*}[!t]
\centering
\begin{promptcard}{Zero-shot / w/o Statistics / w/o Comparison / Tri-level Reasoning Prompt Template}
{\ttfamily\small
You are a \textbf{<DATASET\_DOMAIN>} classifier/estimator.\\
Based on the following \textbf{<ENTITY>} features, predict \textbf{<TARGET\_DESCRIPTION>}.\\
Please explain your reasoning.\\
At the very end of your reasoning, on a separate line, output exactly:\\
Prediction is <value>.\\
where <value> is the \textbf{<PREDICTED\_TARGET\_VALUE\_DESCRIPTION>}.\\
Do not add any other text after this line.\\
\textbf{[If applicable: w/o Statistics]}\\
You are also provided with similar historical cases to assist your reasoning.\\
\textbf{[If applicable: w/o Comparison]}\\
You are also provided with dataset statistics to assist your reasoning.\\
\textbf{[If applicable: Tri-level Reasoning]}\\
You are also provided with dataset statistics and similar historical cases to assist your reasoning.
}
\end{promptcard}
\caption{Prompt template used for Zero-shot / w/o Statistics / w/o Comparison / Tri-level Reasoning.}
\label{tab:prompt-zero-abl-tlr}
\end{table*}

\begin{table*}[t]
\centering
\begin{promptcard}{Standard FT Corpus Construction Prompt Template}
{\ttfamily\small
You are a \textbf{<DATASET\_DOMAIN>} classifier/estimator.\\
Based on the following \textbf{<ENTITY>} features, explain why THE GIVEN PREDICTION LABEL for \textbf{<TARGET\_DESCRIPTION>} is plausible. \textbf{<LABEL\_SEMANTICS\_DESCRIPTION>}\\
At the very end of your reasoning, on a separate line, output exactly:\\
Prediction is <value>.\\
Do not add any other text after this line.
}
\end{promptcard}
\caption{Prompt template for standard and Evidence-Augmented distillation corpus construction.}
\label{tab:prompt-ft-corpus}
\end{table*}

\begin{table*}[t]
\centering
\begin{promptcard}{TLRD Corpus Construction Prompt Template}
{\ttfamily\small
You are a \textbf{<DATASET\_DOMAIN>} classifier/estimator.\\
Based on the following \textbf{<ENTITY>} features, conduct in-depth qualitative analysis to explain why \textbf{THE GIVEN PREDICTION LABEL} for \textbf{<TARGET\_DESCRIPTION>} is plausible.\\
You are also provided with \textbf{DATASET STATISTICS} and \textbf{SIMILAR HISTORICAL CASES} to assist your reasoning.\\
Please explain your reasoning in the following way.\\

Reasoning format:\\
(1). Self feature-based reasoning:\\
- Analyze each feature from ``Current Sample Features'' strictly using domain logic and common sense.\\
Execution Rules:\\
- Briefly explain your intuition, and state its directional impact on the target.\\
- Focus on the self feature logical connection.\\
Prohibitions:\\
- DO NOT use any dataset-level quantities (e.g., mean, percentile) from ``Dataset Statistics'' or similar cases in this step.\\

(2). Global statistics-based reasoning:\\
- Contextualize the sample by comparing its key feature values against the provided dataset statistics for EACH class.\\
- Prioritize those with the clearest class separation or most extreme placement for this sample.\\
- For each feature, explicitly cite the relevant statistics (quote the numbers) when stating where the sample falls relative to the stats (e.g., closer to which class mean/median, or which percentile), highlight deviations/alignments between the sample and the statistical benchmarks, then explain the directional implication.\\

(3). Similar-cases reasoning:\\
Use the provided similar historical cases to extract feature-combination alignment patterns and critical deviations, to further strengthen and validate your prediction logic.\\
Execution Rules:\\
- Extract consistent feature-combination alignment patterns (e.g., [A + B + C] matches).\\
- Identify critical feature deviations (e.g., [D] differs).\\
- For every pattern or deviation mentioned, explicitly cite the current sample's corresponding features.\\
- Clearly explain how the current sample's features align with patterns or diverge from deviations.\\
Prohibitions:\\
- Do NOT reference similar historical cases using numeric identifiers (IDs, order/rank, similarity scores).\\
- Do NOT use the words ``Example'', ``record'', or ``case''.\\
- Structure your response in two sections:\\
1. <Patterns>: List all consistent feature-combination alignment patterns, each explicitly linked to the current sample's matching features.\\
2. <Deviations>: List the most critical feature deviations, each explicitly linked to the current sample's differing features.\\

(4). Final Resolution:\\
Weigh the evidence from the above steps. If there are conflicts, explicitly explain how you resolve them.\\
Use a layered, step-by-step reasoning chain: start from feature logic, then refine with statistics, then validate with similar historical cases.\\

At the very end of your reasoning, on a separate line, output exactly:\\
Prediction is <value>.\\
where <value> is the \textbf{<PREDICTED\_TARGET\_VALUE\_DESCRIPTION>}.\\
Do not add any other text after this line.
}
\end{promptcard}
\caption{Prompt template for TLRD corpus construction.}
\label{tab:prompt-tlrd-corpus}
\end{table*}
\clearpage
\begin{table*}[t]
\centering
\begin{promptcard}{TLRD Model Inference Prompt Template}
{\ttfamily\small
You are a \textbf{<DATASET\_DOMAIN>} classifier/estimator.\\
Based on the following \textbf{<ENTITY>} features, conduct in-depth qualitative analysis to predict \textbf{<TARGET\_DESCRIPTION>}.\\
You should use \textbf{DATASET STATISTICS} and \textbf{SIMILAR HISTORICAL CASES} to assist your reasoning.\\
Please explain your reasoning in the following way.\\

Reasoning format:\\
(1). Self feature-based reasoning:\\
(2). Global statistics-based reasoning:\\
(3). Similar-cases reasoning:\\
(4). Final Resolution:\\

At the very end of your reasoning, on a separate line, output exactly:\\
Prediction is <value>.\\
where <value> is the \textbf{<PREDICTED\_TARGET\_VALUE\_DESCRIPTION>}.\\
Do not add any other text after this line.
}
\end{promptcard}
\caption{Prompt template for TLRD model inference.}
\label{tab:prompt-tlrd-infer}
\end{table*}

\clearpage

\section{Design and Analyses on Hallucination, Consistency, and Human Annotation}
\label{analyse of usefulness}
Given the inherent hallucination risk of LLMs, we explicitly design code-level safeguards, and we further conduct quantitative experiments to assess whether student-generated rationales are evidence-grounded and prediction-consistent. However, hallucination cannot be fully eliminated in current LLMs, and it remains an inherent limitation of LLM-based generation rather than a risk specific to TLRD. Therefore, our design and analyses should be understood as efforts to improve and verify the practical reliability of TLRD, rather than as a claim that hallucination is completely solved or absent. We further conduct a small-scale human study to assess whether the generated rationales are understandable, useful, and well structured from a decision-support perspective.

\subsection{Corpus Construction Safeguards}

\paragraph{Corpus consistency control.}
In the corpus-construction prompt, we state at the beginning that the teacher should explain why \textit{the given prediction label} is plausible. However, in the subsequent rationale-generation steps, we do not repeatedly expose or emphasize the label. This design reduces the chance that the teacher simply rationalizes the label in an unconstrained way. In addition, to prevent data leakage and the hallucination of specific training records during student fine-tuning, the prompt imposes explicit generation constraints: the teacher is forbidden from echoing similarity scores, numeric identifiers, or rank orders in its final output. We also apply a rerun mechanism: if the teacher's final generated prediction does not match the given label, we discard the output and rerun the sample.

\paragraph{Corpus quality control.}
In our quality-control process, we conduct an abnormal-length check on generated rationales. Specifically, we use the IQR rule to identify outputs above the upper fence and manually inspect these long-rationale outliers. We find that about 1\% of the generated rationales contain relatively weak and verbose reasoning. In our practical pipeline, we automatically drop these abnormal generations before fine-tuning.

\paragraph{Evidence support.}
During preliminary validation and optimization, we conducted a small-scale review to assess the usefulness of the selected evidence and to examine whether it provided reasonable support for the generated rationale. However, in complex real-world tabular datasets, especially \textsc{Home Credit}, the decision boundary can be intrinsically noisy. The dataset contains more than 300K loan-application samples, and even strong tabular models achieve relatively low F1 scores on this task. According to our observation, in such datasets, highly similar samples, even with similarity scores as high as 0.99, can still have different labels. Therefore, we do not claim that every generated rationale identifies a uniquely decisive feature pattern. Instead, TLRD aims to provide a general framework for helping LLMs better understand and reason over tabular data by grounding rationales in instance-level, dataset-level, and comparison-level evidence.

\subsection{Quantitative Analysis of Student Outputs}
We conduct quantitative experiments to assess whether student-generated rationales are evidence-grounded and prediction-consistent. Experiments are conducted on four classification datasets by randomly sampling 100 data points from the validation set of each dataset. 
We exclude regression datasets because their numerical outputs can fluctuate substantially, making this evaluation protocol inapplicable to regression tasks. Specifically, we define the following four metrics: Prediction Stability, Prediction--Rationale Consistency, Cited-Number Error, and Similar-Case Support.

\noindent\textbf{Prediction Stability.}
Adapted from SelfCheckGPT for hallucination detection \citep{manakul-etal-2023-selfcheckgpt}, we build a hallucination-detection metric tailored to our tabular prediction setting. This can be explained by the probability-distribution mechanism of LLM generation. When a model is confident about a factual statement, the probability mass of the correct answer tends to dominate the output distribution. For example, for the question ``What is the capital of the United States?'', the model is expected to consistently generate ``Washington, D.C.'' even when moderate randomness is introduced during decoding, such as using a higher temperature or increasing the top-$p$ threshold to enlarge the candidate sampling pool. In our setting, this idea corresponds to measuring the consistency rate of predictions for the same sample under different temperature and top-$p$ configurations.

For each test instance, we generate $R=6$ outputs under different decoding configurations, using two temperature values $\{0.1, 0.2\}$ and three top-$p$ values $\{0.9, 0.95, 1.0\}$. These settings introduce mild decoding perturbations by varying both sampling temperature and nucleus sampling thresholds, while avoiding overly random generation. We compute prediction stability as the fraction of repeated outputs that agree with the majority prediction:
\[
\mathrm{PredStab}(x)=\frac{1}{R}\max_y \sum_{r=1}^{R}\mathbf{1}[\hat{y}^{(r)}=y],
\]
where $\hat{y}^{(r)}$ denotes the final prediction in the $r$-th generated output. A higher score indicates that the model makes more stable decisions under decoding variations. Since $R=6$ yields a discrete stability score, we regard an instance as prediction-stable if at least five out of six generations produce the same final prediction, i.e., $\mathrm{PredStab}(x)\geq 5/6 \approx 0.833$. This threshold allows one minor sampling-induced deviation while filtering out cases where the final prediction changes under multiple decoding configurations.

\begin{table*}[!t]
\centering
\small
\setlength{\tabcolsep}{5pt}
\renewcommand{\arraystretch}{0.9}
\begin{tabular*}{0.9\textwidth}{@{\extracolsep{\fill}}llcccc@{}}
\toprule
\textbf{Dataset} 
& \textbf{Model} 
& \textbf{Stab. $\uparrow$} 
& \textbf{Cons. $\uparrow$} 
& \textbf{Cite Err. $\downarrow$} 
& \textbf{Case $\uparrow$} \\
\midrule

\multirow{3}{*}{\textsc{Adult}}
& \textit{Gemma~3-12B}  & $0.973$ & 100\% & 1/4326 & 97\% \\
& \textit{Qwen3-8B}     & $0.965$ & 99\%  & 0/4598 & 96\% \\
& \textit{Llama~3.1-8B} & $0.935$ & 97\%  & 1/4441 & 98\% \\

\midrule
\multirow{3}{*}{\textsc{Home Credit}}
& \textit{Gemma~3-12B}  & $0.953$ & 100\% & 0/5925 & 98\% \\
& \textit{Qwen3-8B}     & $0.943$ & 100\% & 1/5979 & 99\% \\
& \textit{Llama~3.1-8B} & $0.938$ & 100\% & 1/6401 & 98\% \\

\midrule
\multirow{3}{*}{\textsc{OkCupid}}
& \textit{Gemma~3-12B}  & $0.917$ & 100\% & 2/3307 & 98\% \\
& \textit{Qwen3-8B}     & $0.910$ & 100\% & 2/3165 & 98\% \\
& \textit{Llama~3.1-8B} & $0.903$ & 100\% & 1/3549 & 96\% \\

\midrule
\multirow{3}{*}{\textsc{Diabetes130US}}
& \textit{Gemma~3-12B}  & $0.905$ & 100\% & 2/9132 & 96\% \\
& \textit{Qwen3-8B}     & $0.913$ & 97\%  & 1/8943 & 93\% \\
& \textit{Llama~3.1-8B} & $0.888$ & 98\%  & 3/9307 & 94\% \\

\bottomrule
\end{tabular*}
\caption{Quantitative analyses of hallucination and consistency on classification datasets. 
Stab. denotes prediction stability averaged over all sampled instances; for each instance, 1.000 means that all repeated generations produce the same prediction, while 0.833 means that five out of six generations agree, i.e., one prediction shift. 
Cons. denotes the percentage of samples that pass the prediction--rationale consistency check; 100\% means that all evaluated samples pass this test. 
Cite Err. denotes cited-number error, reported as the number of incorrect cited numbers over the total number of cited numbers. 
Case denotes the percentage of samples whose comparison-level rationales are supported by history cases. 
The selected supervisor is \textit{GPT-OSS-120B}.}
\label{tab:hallucination_consistency_analysis}
\end{table*}
\noindent\textbf{Prediction--Rationale Consistency.}
Prior studies have used simulatability as a faithfulness-oriented criterion for evaluating natural-language explanations, where a useful and faithful explanation should help an observer infer the model's output
\citep{hase-etal-2020-leakage,pan-etal-2025-graphnarrator}. 
Inspired by this idea, we evaluate whether the generated rationale itself provides sufficient support for the model's final prediction. 
For each output, we remove the final resolution statement and the explicit prediction line to avoid direct answer leakage, leaving only the intermediate reasoning process as input to the verifier. 
We then provide this reasoning text to an independent verifier model, GPT-5.5, and ask it to infer the most supported label from the rationale alone. 
If the verifier-inferred label matches the model's original final prediction, we count the output as prediction--rationale consistent. 
This metric measures whether the explanation is aligned with and supportive of the final decision, rather than merely appearing alongside it.

\noindent\textbf{Cited-Number Error.}
We check whether the numerical values cited in the rationales are correct, including both dataset-level statistics and current-sample feature values. 
For dataset-level evidence, we perform tool-assisted verification by giving the verifier access to the precomputed statistics and checking whether each cited statistic matches the corresponding value in the statistics table. 
Although less likely, the model may also misquote values from the current input sample; therefore, we additionally verify whether cited feature values in the rationale are consistent with the actual input features. 
We report cited-number errors as the number of incorrect cited numbers over the total number of cited numbers. 
This metric directly measures numerical hallucination in cited evidence.

\noindent\textbf{Similar-Case Support Rate.}
Finally, we evaluate whether comparison-level evidence is grounded in the retrieved similar cases. For rationales that mention similar historical samples or comparison-level patterns, we provide the verifier with access to the corresponding retrieved training-set references. For a sample, if the cited similar-case evidence can be found in the retrieved context, we count it as supported; otherwise, we count it as unsupported. We report the support rate over all 100 samples.

\paragraph{Experimental Results.}

Table~\ref{tab:hallucination_consistency_analysis} shows that TLRD-generated rationales are generally stable and evidence-grounded. 
Across all classification datasets and student models, prediction stability remains high, suggesting that TLRD enables student models to internalize evidence-grounded reasoning patterns rather than relying on unstable or arbitrary generation.
Prediction--rationale consistency is close to perfect in most cases, indicating that the rationales usually support the model's own predictions. 
Numerical hallucination is rare, with only a few incorrect cited numbers among thousands of cited values. 
Importantly, these errors are occasional rather than systematic: a statistic that is incorrectly cited in one rationale is not consistently cited incorrectly whenever it appears in other rationales. 
This suggests that the errors are more likely caused by the inherent limitations of LLM generation rather than by a systematic flaw in the proposed method. 
In addition, the similar-case support rate remains high across datasets, showing that comparison-level claims are usually grounded in the retrieved examples. 
Overall, these results suggest that TLRD produces rationales that are not only readable, but also stable, internally consistent.

\subsection{Human Evaluation.}

We conduct a small-scale human study to evaluate the usability of generated rationales. 
The study involves five annotators, all of whom have a master's-level or higher academic background in computer science. Evaluations are conducted on four classification datasets by randomly sampling 100 data points from the validation set of each dataset. We focus on classification tasks, where discrete decisions provide a clear basis for judging whether a rationale supports the predicted outcome. Regression rationale evaluation is less directly comparable because continuous-valued predictions require task-specific tolerance ranges, numerical calibration criteria, and stronger domain expertise. 

\paragraph{Evaluation settings.}
Annotators rate each rationale on a 5-point scale for each dimension, allowing half-point increments:
Understandability, Decision-Support Utility, and Structuredness. 
\begin{itemize}[leftmargin=*,nosep]
    \item \textbf{Understandability} measures whether the rationale is easy to read and understand. 

    \item \textbf{Decision-Support Utility} measures whether the rationale provides useful insights for understanding or verifying the model decision in a practical decision-making setting. 

    \item \textbf{Structuredness} measures whether the rationale is well organized and presents the reasoning in a clear and coherent structure.
\end{itemize}

\noindent\textbf{Rating scale.}
\begin{promptbox}
\small
\begin{itemize}[leftmargin=*,nosep]
    \item \textbf{1}: Very poor quality, where the rationale clearly performs poorly in the evaluated dimension.
    \item \textbf{2}: Poor quality, where the rationale has major weaknesses but may still contain some useful elements.
    \item \textbf{3}: Acceptable quality, where the rationale partially satisfies the evaluated dimension but has clear limitations.
    \item \textbf{4}: Good quality, where the rationale mostly satisfies the evaluated dimension with only minor weaknesses.
    \item \textbf{5}: Excellent quality, where the rationale fully satisfies the evaluated dimension.
\end{itemize}
\end{promptbox}

\begin{table}[!t]
\centering
\setlength{\tabcolsep}{2.0pt}
\renewcommand{\arraystretch}{0.85}
\begin{tabular*}
{\columnwidth}{@{\extracolsep{\fill}}llccc@{}}
\toprule
\textbf{Data} 
& \textbf{Method} 
& \textbf{Und. $\uparrow$} 
& \textbf{Util. $\uparrow$} 
& \textbf{Struc. $\uparrow$} \\
\midrule

\multirow{3}{*}{\textsc{Adult}}
& \textit{Base}          & 3.7 & 3.0 & 3.2 \\
& \textit{Std. Distill.} & 3.9 & 3.6 & 3.8 \\
& \textit{TLRD}          & \textbf{4.3} & \textbf{4.4} & \textbf{4.8} \\

\midrule
\multirow{3}{*}{\textsc{HC}}
& \textit{Base}          & 3.2 & 2.4 & 2.6 \\
& \textit{Std. Distill.} & 3.8 & 3.6 & 3.5 \\
& \textit{TLRD}          & \textbf{4.2} & \textbf{4.3} & \textbf{4.8} \\

\midrule
\multirow{3}{*}{\textsc{OK}}
& \textit{Base}          & 3.1 & 2.6 & 2.4 \\
& \textit{Std. Distill.} & 3.7 & 3.6 & 3.6 \\
& \textit{TLRD}          & \textbf{4.1} & \textbf{4.2} & \textbf{4.6} \\

\midrule
\multirow{3}{*}{\textsc{D130}}
& \textit{Base}          & 3.2 & 2.9 & 3.0 \\
& \textit{Std. Distill.} & 4.0 & 3.7 & 4.0 \\
& \textit{TLRD}          & \textbf{4.4} & \textbf{4.6} & \textbf{4.8} \\

\bottomrule
\end{tabular*}
\caption{Human evaluation results on classification datasets using \textit{Gemma~3-12B} as the backbone and \textit{GPT-OSS-120B} as the teacher model. 
\textit{Base}, \textit{Std. Distill.}, and \textit{TLRD} denote the non-finetuned model, standard distillation, and our method, respectively. 
Und., Util., and Struc. denote understandability, decision-support utility, and structuredness.}
\label{tab:human_eval}
\end{table}
\paragraph{Evaluation Results.}
Table~\ref{tab:human_eval} shows that standard distillation improves over the non-finetuned base model in most cases, confirming the benefit of rationale supervision. 
TLRD further achieves the best scores across all datasets and criteria. This indicates that tri-level rationale distillation produces more readable, useful, and structured explanations for tabular decision support.

\begin{table*}[!t]
\setlength{\tabcolsep}{6pt}
\centering
\resizebox{\textwidth}{!}{
\begin{tabular}{lcccccc}
\toprule
\textbf{Model}
& \textbf{Adult}
& \textbf{Home Credit}
& \textbf{Diabetes130US}
& \textbf{OkCupid}
& \textbf{California}
& \textbf{Diamonds} \\
& \textbf{(F1 $\uparrow$)}
& \textbf{(F1 $\uparrow$)}
& \textbf{(Macro-F1 $\uparrow$)}
& \textbf{(Macro-F1 $\uparrow$)}
& \textbf{(RMSE $\downarrow$)}
& \textbf{(RMSE $\downarrow$)} \\
\midrule

\textbf{Llama 3.1-8B (Zero-shot)}
& $51.3$ & $15.5$ & $19.7$ & $46.0$ & $3.02 \times 10^{7}$ & $3.64 \times 10^{8}$ \\
\hspace{1em}w/o Statistics
& $58.9$ & $17.9$ & $29.1$ & $50.2$ & $4.38 \times 10^{6}$ & $4522.6$ \\
\hspace{1em}w/o Comparison
& $58.6$ & $15.6$ & $21.3$ & $40.2$ & $1.25 \times 10^{7}$ & $7315.3$ \\
\hspace{1em}Tri-level Reasoning
& $63.9$ & $17.8$ & $27.1$ & $51.5$ & $1.12 \times 10^{7}$ & $2193.4$ \\
\hspace{1em}Standard Distillation
& $62.5$ & $19.9$ & $25.3$ & $46.8$ & $91399.0$ & $1201.7$ \\
\hspace{1em}Evidence-Augmented Distillation
& $63.1$ & $23.7$ & $24.2$ & $47.4$ & $92592.7$ & $1094.9$ \\
\hspace{1em}TLRD
& $64.7$ & $25.1$ & $34.9$ & $54.2$ & $86572.7$ & $1531.9$ \\

\textbf{Qwen3-8B (Zero-shot)}
& $57.9$ & $20.4$ & $6.1$ & $53.6$ & $175459.5$ & $2280.9$ \\
\hspace{1em}w/o Statistics
& $63.6$ & $18.5$ & $36.3$ & $53.3$ & $65895.3$ & $736.6$ \\
\hspace{1em}w/o Comparison
& $61.4$ & $22.3$ & $29.5$ & $49.9$ & $88828.4$ & $1783.5$ \\
\hspace{1em}Tri-level Reasoning
& $65.5$ & $26.0$ & $36.6$ & $54.0$ & $65660.6$ & $691.0$ \\
\hspace{1em}Standard Distillation
& $61.4$ & $23.7$ & $28.0$ & $51.9$ & $79511.6$ & $1317.5$ \\
\hspace{1em}Evidence-Augmented Distillation
& $53.1$ & $24.0$ & $33.7$ & $47.1$ & $82779.9$ & $1294.7$ \\
\hspace{1em}TLRD
& $65.7$ & $26.3$ & $33.4$ & $55.6$ & $86341.3$ & $1316.6$ \\

\textbf{Gemma 3-12B (Zero-shot)}
& $55.1$ & $15.0$ & $9.2$ & $50.3$ & $568304.3$ & $2589.5$ \\
\hspace{1em}w/o Statistics
& $65.3$ & $16.1$ & $32.6$ & $52.1$ & $68114.6$ & $800.0$ \\
\hspace{1em}w/o Comparison
& $61.1$ & $16.1$ & $24.1$ & $44.2$ & $118604.5$ & $2297.6$ \\
\hspace{1em}Tri-level Reasoning
& $63.1$ & $18.8$ & $33.2$ & $51.1$ & $70633.5$ & $1044.9$ \\
\hspace{1em}Standard Distillation
& $63.2$ & $20.0$ & $27.5$ & $44.8$ & $80762.5$ & $1152.1$ \\
\hspace{1em}Evidence-Augmented Distillation
& $57.7$ & $21.4$ & $34.2$ & $50.3$ & $81151.4$ & $872.1$ \\
\hspace{1em}TLRD
& $63.4$ & $25.7$ & $34.5$ & $56.1$ & $85646.6$ & $945.9$ \\

\textbf{GPT-OSS-20B (Zero-shot)}
& $57.9$ & $16.6$ & $14.2$ & $49.4$ & $201829.8$ & $2477.1$ \\
\hspace{1em}w/o Statistics
& $63.4$ & $15.9$ & $36.7$ & $51.6$ & $67616.8$ & $706.0$ \\
\hspace{1em}w/o Comparison
& $64.9$ & $21.8$ & $36.4$ & $52.9$ & $105755.3$ & $1691.2$ \\
\hspace{1em}Tri-level Reasoning
& $66.9$ & $24.3$ & $36.0$ & $54.5$ & $75992.7$ & $705.1$ \\

\textbf{Qwen3-Next-80B (Zero-shot)}
& $61.9$ & $15.5$ & $8.9$ & $44.5$ & $333534.9$ & $1289.5$ \\
\hspace{1em}w/o Statistics
& $62.2$ & $20.6$ & $34.4$ & $51.9$ & $68405.7$ & $730.6$ \\
\hspace{1em}w/o Comparison
& $62.5$ & $18.7$ & $28.6$ & $50.8$ & $98311.9$ & $1565.6$ \\
\hspace{1em}Tri-level Reasoning
& $63.5$ & $23.5$ & $30.4$ & $52.6$ & $73383.1$ & $698.4$ \\

\textbf{GPT-OSS-120B (Zero-shot)}
& $58.2$ & $16.2$ & $21.1$ & $50.4$ & $114749.6$ & $2632.1$ \\
\hspace{1em}w/o Statistics
& $64.3$ & $19.7$ & $37.4$ & $52.0$ & $61736.4$ & $653.1$ \\
\hspace{1em}w/o Comparison
& $65.0$ & $23.6$ & $38.0$ & $52.6$ & $93657.9$ & $1431.0$ \\
\hspace{1em}Tri-level Reasoning
& $66.6$ & $25.2$ & $37.3$ & $55.0$ & $62906.6$ & $649.7$ \\

\bottomrule
\end{tabular}
}
\caption{Ablation study of tri-level reasoning and distillation variants across classification and regression datasets.}
\label{tab:ablation_tri_level_reasoning}
\end{table*}

\section{Ablation Study}
\label{sec:appendix_ablation}
Before presenting the experiments, we first briefly outline the design motivation of TLRD to help readers better understand our ablation studies. In the domain of LLMs for tabular data, base models typically underperform in zero-shot or few-shot settings. While fine-tuning on labels is a natural solution, it introduces the critical problem of explanation collapse. Therefore, we aim to address these challenges via knowledge distillation from a teacher model. 

TLRD provides the teacher with extra evidence and uses a designed schema to guide structured rationale generation. \textbf{For the ablation of extra evidence, we conduct the ablation experiments of Tri-level Reasoning}, which removes dataset-level statistics or comparison-level retrieval, showing that the full tri-level evidence is generally the most robust. Then it is natural to ask whether we can inject such evidence into a compact student model through fine-tuning, and that's why we construct the Tri-Level Rationale Schema. The corpus construction process of \textbf{TLRD can be viewed as extending standard distillation with two additional factors: additional evidence and the designed schema}. Table~\ref{tab:rq3_standard_vs_tlrd} of our paper compares standard rationale distillation with TLRD and shows that TLRD achieves better performance.  \textbf{To further disentangle these two factors, we add a new distillation ablation} where the teacher receives the same additional evidence from dataset-level statistics and comparison-level retrieval as TLRD, but is not constrained by the designed schema. We denote this variant as \textbf{Evidence-Augmented Distillation}.

\subsection{Ablation of Extra Evidence}
Table~\ref{tab:ablation_tri_level_reasoning} shows that both partial variants, \textit{w/o Statistics} and \textit{w/o Comparison}, can already improve LLM prediction performance, suggesting that comparison-level context and dataset-level statistics each serve as useful standalone evidence sources. Across classification tasks, the combination of evidence generally delivers the best results, making it the most robust configuration overall. These results indicate that the two components are generally complementary, with their integration usually leading to the strongest performance. We also provide a more detailed analysis in Section~\ref{sec:further_analysis} by examining the outputs of different model variants, hoping to offer useful insights for interested readers and future researchers.

\subsection{Ablation of Distillation}

Table~\ref{tab:ablation_tri_level_reasoning} disentangles the role of extra evidence and the designed schema. Evidence-Augmented Distillation shows that providing the teacher with dataset statistics and retrieved neighbors can already help the student acquire dataset-specific tabular knowledge. However, this improvement is unstable and can even fall below standard distillation, indicating that extra evidence alone does not guarantee useful supervision. Under this setting, the teacher generates explanations such as: ``The most similar historical cases (similarity $\geq$ 0.98) are overwhelmingly labeled non-STEM, reinforcing that this sample falls into the non-STEM class.'' Such explanations are low-information and potentially noisy supervision signals. In contrast, under the Tri-Level Rationale Schema, we explicitly prohibit referencing similarity scores and require the model to extract feature-combination patterns and critical deviations from similar cases. This encourages the teacher to generate more structured, evidence-grounded, and informative rationales, which can provide higher-quality supervision.

For regression, the results are more mixed. Regression requires fine-grained numerical calibration rather than discrete decision-boundary learning, so retrieved examples may sometimes provide direct numerical anchors that are easier to exploit in a less constrained format. Thus, while TLRD remains effective compared with zero-shot reasoning, compressing precise quantitative mappings into structured rationales and student parameters is still more challenging. This suggests that regression may benefit from additional calibration-oriented schema designs.

\subsection{Further Analysis}
\label{sec:further_analysis}
Beyond these direct quantitative results, we further conduct a more in-depth analysis by examining dataset characteristics and qualitative patterns in model outputs, hoping to bring useful insight for future researchers.

A notable overall trend is that, on GPT-OSS-20B and GPT-OSS-120B, \textit{w/o Comparison} performs better than \textit{w/o Statistics} in nearly all cases, indicating that these models make more effective use of dataset-level summaries alone. This advantage is closely related to their strong thinking ability. Compared with other models, they analyze feature-level statistics more precisely during the thinking process, which allows them to better understand statistical signals and reason from them more effectively. By contrast, for most other backbones, comparison-level context is often more beneficial than statistics alone, suggesting that local case-based evidence is easier to exploit than abstract global summaries. The effectiveness of comparison-level context, however, is highly constrained by retrieval quality. On \textsc{Home Credit}, severe class imbalance makes it difficult to construct stable and sufficiently balanced local neighborhoods, which substantially weakens the \textit{w/o Statistics} variant. In contrast, on \textsc{Diabetes130US}, comparison-level reasoning remains highly informative because the dataset contains many code-like categorical features, such as diagnoses and medications, for which similar patient records provide meaningful local evidence. Moreover, dataset-level statistics capture global marginal trends, while retrieved comparisons emphasize local patterns around the query instance. When these signals focus on different aspects of the data, combining them may not provide additional gains, especially if one source already offers the dominant predictive cue.

For regression, the results are more mixed. 
Retrieved neighbors may provide useful numerical anchors through concrete target values, helping with scale calibration. 
However, dataset-level statistics can still help when global ranges or categorical structures are informative. 
This suggests that regression may require more calibration-oriented designs, rather than directly relying on the same schema used for classification.

\begin{table*}[!h]
\centering
\small
\setlength{\tabcolsep}{5pt}
\begin{tabular}{llccc}
\toprule
\textbf{Backbone} & \textbf{Method} & \textbf{Home Credit} & \textbf{OkCupid} & \textbf{Diamonds} \\
& & \textbf{Overhead (ms)} & \textbf{Overhead (ms)} & \textbf{Overhead (ms)} \\
\midrule

\multirow{5}{*}{\textit{Llama3-8B}}
& Zero-shot           & 0.00   & 0.00  & 0.00  \\
& w/o Statistics      & 246.56 & 28.53 & 21.96 \\
& w/o Comparison      & 29.38  & 4.16  & 1.61  \\
& Tri-level Reasoning & 263.79 & 35.16 & 22.99 \\
& TLRD(\textit{GPT-OSS-120B}) & 0.00 & 0.00 & 0.00 \\
\midrule

\multirow{5}{*}{\textit{Gemma-3-12B}}
& Zero-shot           & 0.00   & 0.00  & 0.00  \\
& w/o Statistics      & 252.12 & 28.83 & 21.66 \\
& w/o Comparison      & 32.47  & 4.72  & 1.27  \\
& Tri-level Reasoning & 284.71 & 34.22 & 23.48 \\
& TLRD(\textit{GPT-OSS-120B}) & 0.00 & 0.00 & 0.00 \\
\midrule

\multirow{5}{*}{\textit{Qwen3-8B}}
& Zero-shot           & 0.00   & 0.00  & 0.00  \\
& w/o Statistics      & 246.34 & 28.70 & 21.94 \\
& w/o Comparison      & 38.10  & 6.22  & 2.11  \\
& Tri-level Reasoning & 273.38 & 33.28 & 22.62 \\
& TLRD(\textit{GPT-OSS-120B}) & 0.00 & 0.00 & 0.00 \\
\bottomrule
\end{tabular}
\caption{Average prompt-construction overhead before the LLM generation call. Overhead is measured in milliseconds and defined as \textit{Comparison Build (ms)} $+$ \textit{Stats Compute (ms/sample)}. Zero-shot and TLRD incur no test-time overhead because they do not require prompt-time statistics computation or retrieval.}
\label{tab:overhead_appendix}
\end{table*}

\section{Task-Specific Optimization Directions}
\label{sec:appendix_task_specific_optimization}

TLRD is designed as a general framework for converting label-only tabular datasets into structured rationale supervision. 
Beyond the default setting used in our main experiments, there are several practical directions that may further improve predictive performance and decision-support utility. 
These include standard optimization choices such as systematic hyperparameter search, checkpoint selection, and full-parameter supervised fine-tuning. TLRD currently uses direct dataset-level statistics, such as class-conditional means and categorical frequencies. A promising extension is to expose higher-order statistical evidence to the teacher model such as the likelihood ratio. For example, conditional likelihood ratios can describe how much additional information a pattern provides after another pattern is already observed, such as whether pattern $B$ remains informative conditioned on pattern $A$. Carefully selecting and presenting such higher-order evidence may help the teacher identify more discriminative feature interactions.
In addition, we explored several task-specific directions that are not included in our experiments but may be useful for future work and for deploying TLRD in specific application scenarios.

\paragraph{Domain-specific reasoning workflow.}
First, even before fine-tuning, the reasoning behavior of LLMs can be improved a little by adapting the prompt to the task-specific decision process while keeping the input features unchanged. 
For example, in \textsc{Home Credit}, the goal is to assess whether an applicant may default on a loan. 
Instead of asking the model to reason in a generic way, we can prompt it to separately analyze \textit{Financial Resilience} and \textit{Behavioral Risk Signals}, which may follow a practical credit-risk assessment workflow. Incorporating this structured approach into rationale construction may yield both better performance and enhanced readability.

\paragraph{Task-specific decision preference.}
Second, the model behavior can be meaningfully affected by explicitly specifying the downstream decision preference. 
For instance, in loan-default detection, missing a true default applicant may be more costly than incorrectly flagging a non-default applicant. 
We find that simply adding an instruction such as ``identifying default is more important than avoiding false alarms'' can noticeably shift the model behavior toward higher recall and lower precision, which may improve F1 under imbalanced settings. 
This observation suggests that LLM-based tabular decision systems can be adjusted according to task-specific cost preferences, rather than treating all prediction errors as equally important.

\begin{table*}[htb]
\centering
\small
\setlength{\tabcolsep}{4.5pt}
\resizebox{\textwidth}{!}{
\begin{tabular}{llcccccc}
\toprule
\textbf{Backbone} & \textbf{Method}
& \multicolumn{2}{c}{\textbf{HC}}
& \multicolumn{2}{c}{\textbf{OK}}
& \multicolumn{2}{c}{\textbf{Diamonds}} \\
\cmidrule(lr){3-4}\cmidrule(lr){5-6}\cmidrule(lr){7-8}
& & \textbf{InputTok} & \textbf{OutTok}
  & \textbf{InputTok} & \textbf{OutTok}
  & \textbf{InputTok} & \textbf{OutTok} \\
\midrule

\multirow{5}{*}{\textit{Llama3-8B}}
& Zero-shot              & 484.15  & 553.67  & 385.68  & 610.40  & 376.36  & 296.90 \\
& w/o Statistics         & 3309.39 & 582.22  & 2646.93 & 369.68  & 1776.80 & 237.79 \\
& w/o Comparison         & 2164.45 & 749.40  & 3599.93 & 725.39  & 2401.88 & 370.74 \\
& Tri-level Reasoning    & 4976.69 & 759.78  & 5848.18 & 361.33  & 3789.32 & 250.30 \\
& TLRD(\textit{GPT-OSS-120B}) & 542.15 & 1930.00 & 442.68 & 1879.42 & 438.36 & 1219.98 \\
\midrule

\multirow{5}{*}{\textit{Gemma-3-12B}}
& Zero-shot              & 541.17  & 708.72  & 392.10  & 787.09  & 376.85  & 645.62 \\
& w/o Statistics         & 4225.43 & 446.93  & 2728.87 & 389.15  & 1936.50 & 346.99 \\
& w/o Comparison         & 2804.11 & 834.65  & 4115.62 & 849.74  & 2893.23 & 487.26 \\
& Tri-level Reasoning    & 6474.37 & 505.82  & 6438.39 & 349.91  & 4438.88 & 506.99 \\
& TLRD(\textit{GPT-OSS-120B}) & 607.17 & 2318.35 & 458.10 & 2154.08 & 446.85 & 1414.09 \\
\midrule

\multirow{5}{*}{\textit{Qwen3-8B}}
& Zero-shot              & 503.08  & 646.39  & 366.33  & 518.79  & 368.85  & 399.23 \\
& w/o Statistics         & 4009.86 & 658.01  & 2706.64 & 701.66  & 1928.50 & 571.38 \\
& w/o Comparison         & 2695.82 & 902.07  & 4014.88 & 1165.16 & 2828.31 & 600.08 \\
& Tri-level Reasoning    & 6189.60 & 934.62  & 6342.19 & 1228.84 & 4374.96 & 701.49 \\
& TLRD(\textit{GPT-OSS-120B}) & 561.08 & 2005.50 & 423.33 & 1971.29 & 430.85 & 1255.31 \\
\bottomrule
\end{tabular}}
\caption{Average input and output token counts for each dataset--model configuration. HC = \textsc{Home Credit}, OK = \textsc{OkCupid}. Token counts are computed using the tokenizer corresponding to each backbone model.}
\label{tab:token_lengths_appendix}
\end{table*}
\section{Inference Efficiency}
\label{sec:appendix_efficiency}

We further report inference-efficiency statistics for different settings. Specifically, we include three quantities: (1) \textbf{overhead}, (2) \textbf{input tokens}, and (3) \textbf{output tokens}. All measurements are conducted with \texttt{vLLM} using a batch size of 1 to simulate an online serving scenario. For each dataset--model configuration, we measure the \textbf{overhead} before the actual LLM generation call, i.e., the time required to construct the final prompt for the query instance, including any auxiliary processing such as assembling retrieved context and dataset statistics when applicable. Overhead is reported in milliseconds and averaged over all evaluated queries. In addition, we report the average \textbf{input length} (number of prompt tokens fed into the model) and the average \textbf{output length} (number of generated tokens in the model response).

Table~\ref{tab:overhead_appendix} reports the average prompt-construction overhead in milliseconds for each dataset--model configuration. In general, the overhead is dominated by comparison-context construction rather than statistics computation, as the \textit{w/o Statistics} setting consistently incurs much higher cost than \textit{w/o Comparison} across all datasets. This gap is especially pronounced on \textsc{Home Credit}, where retrieval-based prompt construction is substantially more expensive than on \textsc{OkCupid} and \textsc{Diamonds}.

Table~\ref{tab:token_lengths_appendix} reports the corresponding average input and output lengths, computed using the tokenizer of each backbone model. We do not separately emphasize end-to-end inference latency here, because in our setting it is largely dominated by generated output length, which is in turn controllable through the length of teacher-generated rationales during corpus construction. We therefore report token statistics mainly to provide a practical view of the prompting cost under different settings. Excluding TLRD, the main pattern is that prompt-augmented inference trades substantially longer inputs for only moderate changes in output length: \textit{w/o Statistics}, \textit{w/o Comparison}, and especially \textit{Tri-level Reasoning} increase the input length by attaching external context at test time, while their generated outputs remain in a similar range or grow only moderately depending on the backbone and dataset. By contrast, TLRD keeps the input length close to zero-shot prompting because it requires no test-time statistics or retrieval, but produces much longer outputs. This is consistent with our design: the student internalizes a richer reasoning style distilled from \textit{GPT-OSS-120B}, shifting more of the reasoning cost from prompt construction to response generation. In practice, this output cost could be reduced when needed by constraining teacher-generated rationales to be more concise.

\section{Label-only Fine-tuning Results}
\label{sec:appendix_label_only_ft}
\begin{table}[!t]
\centering
\small
\setlength{\tabcolsep}{2.2pt}
\begin{tabular*}{\columnwidth}{@{\extracolsep{\fill}}lccc}
\toprule
Method & HC $\uparrow$ & OK $\uparrow$ & Dia. $\downarrow$ \\
\midrule
Llama~3.1-8B Label-only FT & 28.3 & 51.3 & 1100.9 \\
Qwen3-8B Label-only FT     & 19.0 & 53.0 & 1682.1 \\
Gemma~3-12B Label-only FT  & 28.1 & 58.7 & 688.7 \\
\bottomrule
\end{tabular*}
\caption{Results of label-only fine-tuning. HC = \textsc{Home Credit}, OK = \textsc{OkCupid}, and Dia. = \textsc{Diamonds}.}
\vspace{-0.1cm}
\label{tab:label_only_ft}
\end{table}

To observe the explanation-collapse phenomenon discussed in prior work \citep{yang2025beyond}, we also conduct label-only fine-tuning experiments. Table~\ref{tab:label_only_ft} reports the prediction outcomes of label-only fine-tuned models when they are evaluated in the same format as fine-tuning, i.e., mapping the input directly to the final prediction label without generating any rationale. Although some results seem competitive, they are accompanied by an almost complete loss of explanation ability.
When we additionally prompt the label-only fine-tuned model to provide explanations, the model often fails to follow the instruction, producing either malformed outputs or still label-only responses. Even when it does follow the explanation instruction, the generated rationales are shallow and come with a large performance drop.
For instance, after being explicitly asked to explain its decision, \textit{Gemma~3-12B} drops to 49.0 Macro-F1 on \textsc{OkCupid}, even below its zero-shot performance. 
In one representative error, the model misclassifies a non-\textsc{stem} job profile as \texttt{student} and explains: ``The profile indicates the education level is a `masters program', which strongly suggests a student. The other features, such as age, sex, location, etc. are not as important for determining the job category in this case.'' 
This explanation is low-information and reveals that the model no longer understands the feature semantics: it unreasonably conflates a master's-level educational background with the student job category.

The experimental observations are consistent with prior work, when optimized solely with label supervision, the model is encouraged to minimize the discriminative objective with the shortest valid output, rather than to preserve free-form explanation ability. 
As a result, LLMs can behave like a discriminative Transformer-style predictor over serialized rows and may achieve performance comparable to some neural tabular models on certain tasks. 
At the same time, it almost completely loses its ability to generate meaningful explanations, and forcing explanation generation can further degrade prediction performance.

\section{Explainability of Tree-Based Methods and TLRD}
\label{sec:appendix_explainability_tree_tlrd}

Tree-based methods such as XGBoost can only provide relatively coarse explanation signals, such as feature importance or attribution scores. However, for complex real-world tabular data, these signals cannot provide case-by-case analysis value. Moreover, modern tree ensembles typically consist of thousands of estimators, making case-by-case path tracing practically valueless.

For example, on \textsc{Home Credit}, XGBoost can produce feature importance such as:
\begin{promptbox}
\small
\centering
\begin{tabular}{lr}
\texttt{ORGANIZATION\_TYPE} & 0.242968 \\
\texttt{OCCUPATION\_TYPE} & 0.077036 \\
\texttt{NAME\_TYPE\_SUITE} & 0.028893 \\
\texttt{NAME\_INCOME\_TYPE} & 0.026812 \\
\texttt{WALLSMATERIAL\_MODE} & 0.026786 \\
\texttt{NAME\_EDUCATION\_TYPE} & 0.025851 \\
\texttt{NAME\_HOUSING\_TYPE} & 0.023091 \\
\texttt{WEEKDAY\_APPR\_PROCESS\_START} & 0.022930 \\
\texttt{NAME\_FAMILY\_STATUS} & 0.017258 \\
\texttt{FONDKAPREMONT\_MODE} & 0.016404 \\
\end{tabular}
\end{promptbox}
Such feature-importance results tell us which columns are generally useful for prediction, but they do not directly explain why a specific applicant is predicted as default or non-default. In contrast, TLRD generates case-specific rationales that connect the current feature values with dataset-level statistical evidence and similar-case contrastive evidence. This allows the explanation to answer questions such as: whether a feature value is statistically unusual, whether it is discriminative between classes, and how the current sample aligns with or deviates from similar historical samples. Below, we show the final resolution from an example TLRD-generated rationale:

\begin{promptbox}
\small
The self-feature analysis flags several high-risk signals: low external scores, missing external scores, short employment tenure, young age, high annuity, and a male laborer with secondary education. 
Global statistics confirm that the applicant's \texttt{EXT\_SOURCE\_2}, age, employment length, and annuity are closer to the default distributions, while the modest goods price and recent ID update provide only minor offsets. 
Although most highly similar historical samples with the same categorical profile have a non-default outcome, the key quantitative deviations, including lower \texttt{EXT\_SOURCE\_2}, younger age, and higher annuity, tilt the balance toward higher default risk. 
Consequently, the overall evidence supports the given prediction of default.
\end{promptbox}

\section{Efficiency Comparison with Tabular Baselines}
\label{sec:appendix_efficiency_comparison}

Prior work has shown that, in practical tabular prediction settings, tree-based methods such as XGBoost and CatBoost remain highly competitive and efficient~\citep{shwartz2022tabular,grinsztajn2022tree}. In Table~\ref{tab:combined_by_backbone}, XGBoost and CatBoost remain among the strongest baselines for pure tabular prediction across most datasets.
We further conduct efficiency experiments and find that this observation still holds in our setting.
All measurements are conducted on a 24GB NVIDIA RTX A5000 GPU. 
For Qwen3-8B TLRD, we use \textit{GPT-OSS-120B} as the teacher model and vLLM for inference. 
The elapsed-time values for Qwen3-8B TLRD report inference time only and do not include fine-tuning time.
In our experiments, fine-tuning the LLM backbones typically takes around 3--6 hours on 4$\times$80GB A100 GPUs, and inference requires at least a 24GB GPU. 
For reference, on \textsc{Home Credit}, a fine-tuned Qwen3-8B model supervised by \textit{GPT-OSS-120B} takes a median of 9.71 seconds to generate one complete rationale.

\begin{table*}[!t]
\centering
\small
\setlength{\tabcolsep}{4pt}
\begin{tabular}{lcccccc}
\toprule
\textbf{Model} 
& \textbf{Adult} 
& \textbf{Home Credit} 
& \textbf{Diabetes} 
& \textbf{OkCupid} 
& \textbf{California} 
& \textbf{Diamonds} \\
\midrule
TabPFN(v2.5)  & 584.10  & 10198.61 & 972.57  & 833.54  & 58.45   & 280.38  \\
TabPFN(v2.0)  & 418.17  & 12852.28 & 1598.24 & 847.72  & 48.45   & 359.95  \\
XGBoost       & 1921.03 & 7108.51  & 3216.24 & 4594.69 & 3106.04 & 800.34  \\
CatBoost      & 1040.01 & 768.20   & 1152.22 & 737.84  & 1578.13 & 714.41  \\
TabM          & 2317.10 & 18271.61 & 3314.52 & 3120.38 & 3333.91 & 6620.48 \\
Qwen3-8B TLRD & 2390.20 & 2740.44 & 2691.00 & 2388.54 & 1455.57 & 2361.72 \\
\bottomrule
\end{tabular}
\caption{Total elapsed wall-clock time in seconds. Lower is better.}
\label{tab:efficiency_runtime}
\end{table*}

\begin{table*}[!t]
\centering
\small
\setlength{\tabcolsep}{4pt}

\begin{tabular}{lcccccc}
\toprule
\textbf{Model} 
& \textbf{Adult} 
& \textbf{Home Credit} 
& \textbf{Diabetes} 
& \textbf{OkCupid} 
& \textbf{California} 
& \textbf{Diamonds} \\
\midrule
TabPFN(v2.5)  & 1.34  & 23.46 & 5.40  & 2.63  & 1.84  & 6.30  \\
TabPFN(v2.0)  & 3.34  & 23.42 & 12.74 & 4.37  & 2.08  & 7.34  \\
XGBoost       & 0.33  & 0.48  & 0.52  & 0.79  & 0.36  & 0.33  \\
CatBoost      & $\sim$22.5 & $\sim$22.5 & $\sim$22.5 & $\sim$22.5 & $\sim$22.5 & $\sim$22.5 \\
TabM          & 0.60  & 1.40  & 1.34  & 1.81  & 1.21  & 1.21  \\
Qwen3-8B TLRD & $\sim$21.3 & $\sim$21.3 & $\sim$21.3 & $\sim$21.3 & $\sim$21.3 & $\sim$21.3 \\
\bottomrule
\end{tabular}
\caption{Peak GPU memory usage in GB. Values marked with $\sim$ are approximate peak allocations.}
\label{tab:efficiency_memory}
\end{table*}

Tables~\ref{tab:efficiency_runtime} and~\ref{tab:efficiency_memory} confirm the strong efficiency advantage of tree-based methods.
XGBoost remains the most resource-efficient baseline in terms of GPU memory usage across all datasets. 
On larger datasets \textsc{Home Credit}, XGBoost and CatBoost also achieve faster execution than TabPFN. 
For CatBoost, the reported GPU memory mainly reflects its default GPU memory reservation behavior. 
By default, CatBoost uses \texttt{gpu\_ram\_part = 0.95}, so it may reserve a large fraction of available GPU memory after initialization. For Qwen3-8B TLRD, the reported GPU memory also reflects the vLLM configuration. We launch vLLM with \texttt{--gpu-memory-utilization 0.9}.
Therefore, the reported value should be interpreted as observed peak allocation under the default configuration, rather than the strictly required model memory.

We emphasize that TLRD is not designed to compete with tree-based methods in high-throughput deployment. 
Instead, TLRD aims to help bridge the gap between deep learning methods and traditional tabular models in tabular decision-making: it improves the ability of LLM-based models to produce competitive predictions while also generating case-specific rationales that offer valuable reference for decision-making.
Therefore, tree-based models and TLRD can be used in a complementary workflow. 
Tree-based models can serve as efficient high-throughput predictors, while TLRD can be applied in latency-tolerant stages to generate case-specific explanations for selected samples. 
These explanations can support downstream workflows such as assisting data scientists in understanding dataset-specific patterns, auditing individual model decisions, and helping machine learning engineers diagnose or tune tree-based models.

\section{LLM Usage}

In compliance with the ACL Rolling Review (ARR) guidelines regarding AI assistance, we hereby clarify the role of Large Language Models (LLMs) in this work. Since LLMs serve as the primary focus of our empirical investigation, all reported metrics and outputs are derived strictly from our direct execution and evaluation of these models. Furthermore, we employed LLMs solely as text-editing utilities to refine our prose and enhance readability. We confirm that no AI tools were involved in formulating the research hypotheses, designing the framework, or interpreting the scientific findings.

\end{document}